\documentclass[acmtog,authorversion]{acmart}

\usepackage{wrapfig} 
\usepackage{nicefrac}
\usepackage{colortbl}  

\renewcommand\footnotetextcopyrightpermission[1]{} 
\begin{document}

%
%
\title{%
Polarimetric Spatio-Temporal Light Transport Probing
} 

\author{Seung-Hwan Baek}
\affiliation{%
  \institution{Princeton University}
  \department{Computer Science}
}

\author{Felix Heide}
\affiliation{%
  \institution{Princeton University}
  \department{Computer Science}
}

\renewcommand{\shortauthors}{Baek, Heide}

\begin{abstract}
Light emitted from a source into a scene can undergo complex interactions with multiple scene surfaces of different material types before being reflected towards a detector. During this transport, every surface reflection and propagation is encoded in the properties of the photons that ultimately reach the detector, including travel time, direction, intensity, wavelength and polarization. Conventional imaging systems capture intensity by integrating over all other dimensions of the incident light into a single quantity, hiding this rich scene information in {these aggregate} measurements. Existing methods are capable of untangling these measurements into their spatial and temporal dimensions, fueling geometric scene understanding tasks. However, examining polarimetric material properties jointly with geometric properties is an open challenge that could enable unprecedented capabilities beyond geometric scene understanding, {allowing for material-dependent scene understanding} and imaging through complex transport, such as macroscopic scattering.

In this work, we close this gap, and propose a computational light transport imaging method that captures the spatially- and temporally-resolved complete polarimetric response of a scene, which encode{s} rich material properties.
Our method hinges on a novel 7D tensor theory of light transport. We discover low-rank structure in the polarimetric tensor dimension and propose a data-driven rotating ellipsometry method that learns to exploit redundancy of polarimetric structure.
We instantiate our theory {with} two imaging prototypes: spatio-polarimetric imaging and coaxial temporal-polarimetric imaging. This allows us, for the first time, to decompose scene light transport into temporal, spatial, and complete polarimetric dimensions that unveil scene properties hidden to conventional methods. We validate the applicability of our method on diverse tasks, including shape reconstruction with subsurface scattering, seeing through scattering {media}, untangling multi-bounce light transport, breaking metamerism with polarization, and spatio-polarimetric decomposition of crystals.
\end{abstract}

 \begin{teaserfigure}
  \centering
   \includegraphics[width=0.99\linewidth]{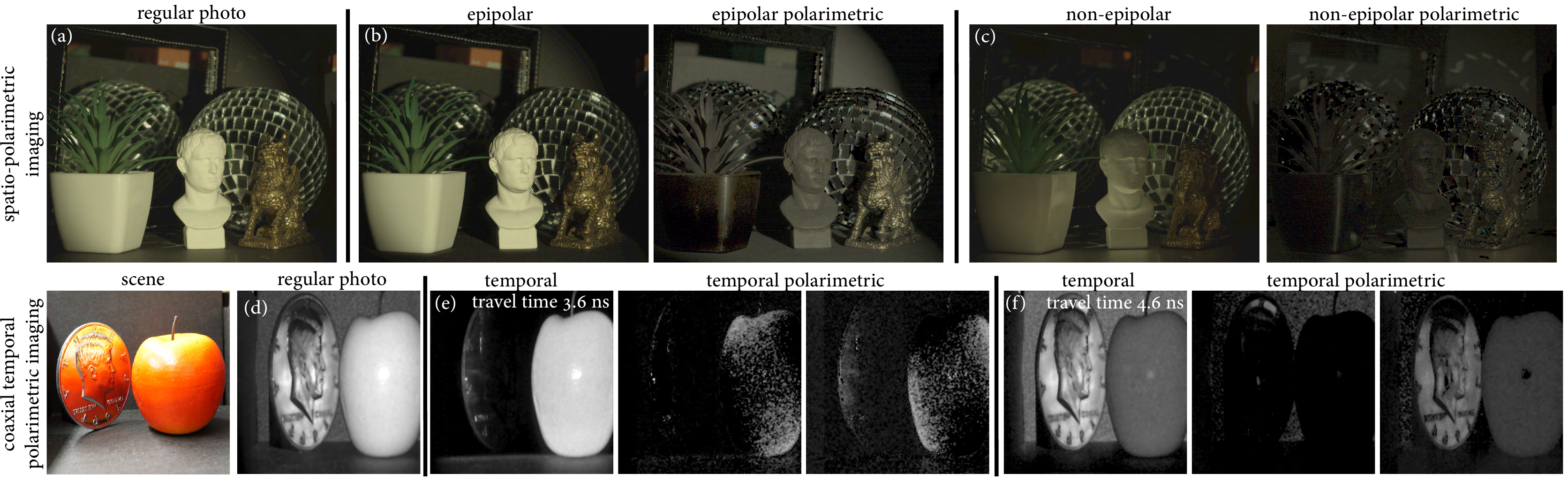}%
   \vspace{-2mm}%
   \caption[]{\label{fig:teaser}%
    We propose a computational light transport {probing method that decomposes transport into full polarization, spatial and temporal dimensions}.
    We model this multi-dimensional light transport as a tensor and analyze low-rank structure in the polarization domain which is exploited by our polarimetric probing method.
     We instantiate our approach with two imaging systems for spatio-polarimetric and coaxial temporal-polarimetric capture.
    (a)\&(d) Conventional intensity imagers integrate incident light intensity over space and time independently of the polarization states of light, losing geometric and material information encoded in the polarimetric transport.
    Capturing polarization-resolved spatial transport components of (b) epipolar and (c) {non-epipolar} dimensions enable {fine-grained decomposition of light transport}.
    Combining temporal and polarimetric dimensions, we separate (e) geometry-dependent reflections and (f) direct/indirect reflections that cannot be resolved in the temporal-only measurements.
    }
\end{teaserfigure}

\maketitle

\section{Introduction}
\label{sec:introduction}
Light transport theory has proven itself to be a powerful tool across fields that rely on how light flows from a source {through a scene} to a detector. {As} photons travel through the scene, they interact with matter on surfaces or {subsurfaces.} {This interaction modifies the photon properties (light direction, travel time, intensity, wavelength, and polarization) and, as such, encodes information about the scene}. These properties have been formalized as plenoptic light dimensions augmented with wave properties~\cite{adelson1991plenoptic,wetzstein2011computational}. Conventional detectors integrate over {all} dimensions of this high-dimensional plenoptic function. While this integration is beneficial in terms of photon efficiency, the individual plenoptic dimensions that encode scene information are lost {during} the conventional measurement process.

To recover this hidden scene information, a large body of work in computer graphics and vision has explored light transport approaches that aim to decompose the integral into a subset of the plenoptic dimensions. Specifically, considerable progress has been made in untangling spatial and temporal dimensions of light transport.
Spatially controlling emitted and detected light allows for path-dependent geometric scene understanding~\cite{o2012primal,o20143d}.
Similarly, advances in fast illumination and detection devices have allowed for temporal light transport capture~\cite{velten2013femto,heide2013low,o2014temporal}.
Joint spatio-temporal decomposition has also been explored recently~\cite{o2014temporal,kotwal2020interferometric}.
While these existing light transport decomposition methods have been successful for geometric scene understanding, {analyzing material properties has been an open challenge.}
Real-world materials are complex and {shape transport in the light-matter interaction}.
Specifically, we observer complex material-dependent scattering effects such as complex reflection, subsurface scattering, backscattering, multiple reflection between surfaces, and refractive-index-dependent reflection.
Decomposing material-dependent light transport effects could allow for scene understanding and reconstruction tasks that exploit material properties, potentially fuelling broad applications across disciplines, including localization in VR/AR, object digitization, seeing through scattering {media}, and material analysis.

Polarization is an often overlooked dimension that may {allow for analysis of material-dependent scattering} across application domains.
Polarization is a wave property of light describing how the electric field of light oscillates along the propagation direction in time.
Previous polarization imaging methods in graphics focus on cross-polarization configurations that use either linear or circular polarization filters in front of a camera and a light source with perpendicular orientation~\cite{ma2007rapid,treibitz2008active,gupta2008controlling,ghosh2010circularly,ghosh2011multiview}.
This approach, however, fundamentally prevents {measurement of} the full polarimetric response of the light transport, excluding polarization states of light such as elliptically polarized light and partially polarized light.
Using linear polarizers at multiple angles as in polarization cameras has been also extensively used in computer vision~\cite{ba2020deep,atkinson2006recovery,kadambi2015polar,tanaka2019enhancing}, however, this configuration cannot acquire circular and elliptical polarization, and {does not utilize} the polarization of the emitted light.
While Baek et al.~\shortcite{baek2018simultaneous} proposed using a pair of rotating linear polarizers for reflectance acquisition, their approach also cannot model elliptically and circularly polarized light.
More recently, goniometry systems have allowed researchers to acquire isotropic polarimetric BRDFs with full polarimetric resolution~\cite{baek2020image}.
Orthogonal to the proposed light transport acquisition method, this method captures spherical homogenous objects and hence is not applicable to light transport analysis in a complex scene.
{In addition, it cannot capture the space and time dimensions.}

In this work, we analyze material- and geometry-dependent scene transport effects by proposing the first spatio-temporal light transport imaging methods with full polarimetric resolution.
The proposed acquisition method allows us to decompose light into components that have a specific travel time, path geometry and polarization state -- allowing for a 7-dimensional representation of light transport covering 2D camera space, 2D illumination space, 1D travel time, {and} 2D polarization change.
To this end, we propose an imaging method that combines ellipsometry, a technique in optics for characterizing polarimetric response, and spatial and temporal light transport acquisition from graphics and vision.
We also introduce a novel tensor-based light transport theory that maintains natural backward compatibility to existing matrix-based light transport models in order to analyze and decompose multi-dimensional light transport.
Given the high dimensionality of the full 7D light transport, directly acquiring the tensor is impractical with existing time-resolved sensors, and for the experimental setup shown in this work would require \emph{6 years per scene}.
Instead, we analyze the formalized 7D light transport and identify low-dimensional embeddings of the polarimetric light transport that we exploit in our acquisition method. However, in contrast to existing method{s} for space and {wavelength}~\cite{o2010optical,saragadam2019krism} that \emph{directly} acquire low-rank transport, \emph{it is infeasible to directly acquire low-rank polarimetric light transport} as we have to solve an additional reconstruction problem, converting the intensity measurements to the Stokes-Mueller domain~\cite{azzam1978photopolarimetric}.
To overcome this challenge, we propose a novel data-driven rotating ellipsometry method that learns to rely on {the discovered low-dimensional} structure in the polarimetric dimension.
We then instantiate our approach with a projector-camera ellipsometric imager and a coaxial SPAD-laser ellipsometric imager, that allow for spatio-polarimetric and coaxial temporal-polarimetric light transport imaging, respectively.
We validate our method in simulation and experimentally using these prototype systems.
We demonstrate novel applications of the recovered spatio-polarimetric and temporal-polarimetric light transport decompositions, including shape reconstruction with subsurface scattering, seeing through scattering {media}, untangling multi-bounce light transport, spatio-polarimetric decomposition of crystals, and breaking metamerism with polarization.

Specifically, we make the following contributions in this work
\begin{itemize}
  \item We propose a computational light transport imaging method for shape and material analysis that jointly captures full polarimetric information along with spatial and temporal transport dimensions, spanning 7-dimensional light transport.\vspace{4pt}
  \item We introduce a novel tensor-based light transport theory that is naturally backward-compatible with existing matrix models.\vspace{4pt}
  \item We analyze the low-rank structure of polarimetric light transport and propose a learned ellipsometry method which implicitly exploits this low-dimensional polarimetric embedding.\vspace{4pt}
  \item We instantiate two experimental prototypes of the proposed method for spatio-polarimetric imaging and coaxial temporal-polarimetric imaging.\vspace{4pt}
  \item We validate the method with five applications, including shape reconstruction with subsurface scattering, seeing through scattering {media}, multi-bounce light transport decomposition, breaking metamerism with polarization, and spatio-polarimetric decomposition of crystals, where the proposed method outperforms existing approaches in \emph{all} experiments.
\end{itemize}

\vspace{10pt}
\paragraph{Overview of Limitations}
While the proposed imaging method makes it possible, for the first time, {to acquire the complete polarization information together with} the spatial and temporal dimensions, each of the two prototype systems only captures partial dimensions of the 7D spatio-temporal polarimetric tensor we analyze in this work.
Although both systems could be combined with a high-resolution SPAD array, as proposed in {Morimoto et al.~\shortcite{Morimoto:20}, pixel counts of existing commodity sensors are limited to a few kilopixels.} We leave the substantial engineering efforts of such a joint high-resolution temporal acquisition approach for future research.
Also, the capture time of the proposed experimental prototypes is currently restricted to sequential acquisition on the order of hours per scene, limiting the proposed method to static scenes.

\vspace{-4mm}
\section{Related Work}
\label{sec:relatedwork}
\begin{figure*}[t]
  \centering
  \includegraphics[width=0.95\linewidth]{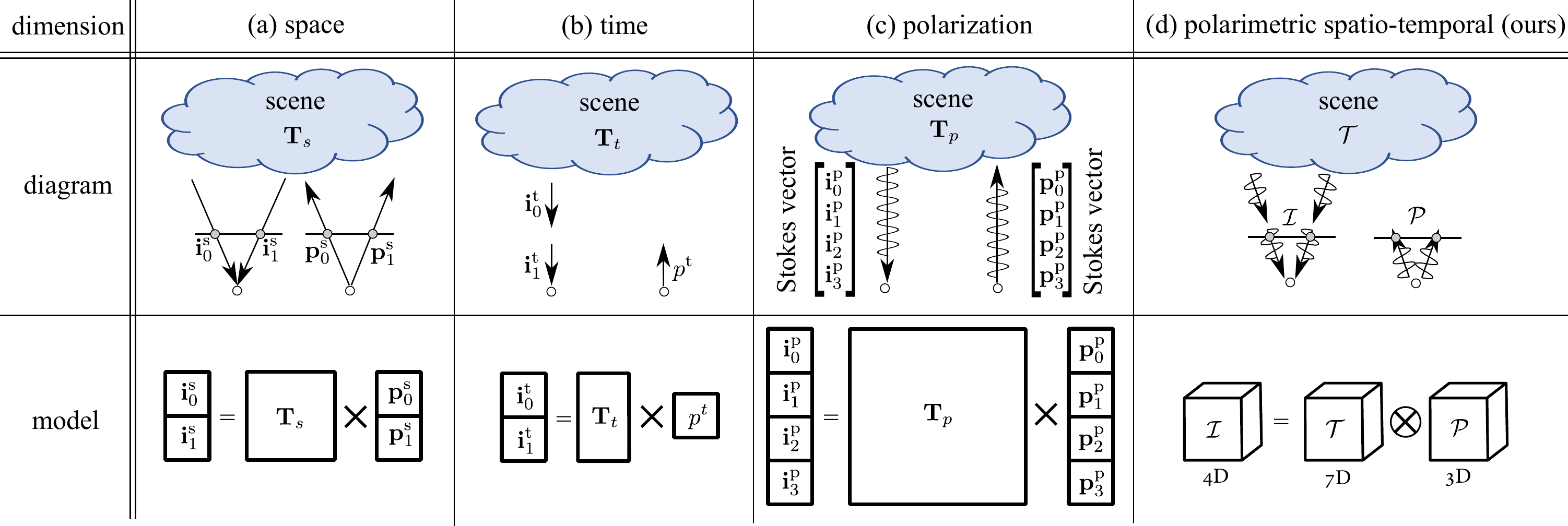}
  \caption{\label{fig:LT_matrix}%
  Previous light transport models for (a) space~\cite{o2012primal}, (b) time~\cite{velten2013femto}, and (c) polarization~\cite{collett2005field} are based on linear modeling as a matrix-vector multiplication. (d) We propose a tensor-based light transport model which simultaneously comprises space, time, and polarization. We model light transport as a 7D tensor  $\mathcal{T}$ covering 2D camera space, 2D illumination space, 1D travel time, and 2D polarization change. This 7D transport is then multiplied {with} a 3D illumination tensor $\mathcal{P}$ covering 2D illumination space and 1D polarization, resulting in a captured 4D tensor $\mathcal{I}$ covering the 2D camera space, 1D polarization, and 1D travel time.
  }
\end{figure*}

\paragraph{Computational Light Transport}
Acquiring and decomposing light transport has been extensively studied in computer graphics and vision.
Early approaches exploit statistics of global and direct reflections to decompose light transport based on the number of light-matter bounces~\cite{nayar2006fast,gupta2011structured}.
Geometric constraints such as stereoscopic epipolar geometry have been also exploited to selectively capture light paths that satisfy the certain geometric constraints~\cite{o2010optical,o2012primal,o20143d,kubo2018acquiring,o2015homogeneous}.
These spatial decomposition methods are typically implemented with a camera and a projector where the projection is masked during the acquisition process.
Another line of research untangles the temporal dimension of light transport~\cite{velten2013femto,wu2014decomposing,o2014temporal,o2017reconstructing,jarabo2017recent}. 
Combined with a fast pulsed laser, direct temporal sampling approaches rely on a streak camera~\cite{velten2013femto} or a single-photon avalanche diode (SPAD)~\cite{o2017reconstructing} to directly measure such temporal scene responses.
An alternative line of work relies on a correlation sensor and amplitude-synchronized continuous illumination to acquire temporal light transport~\cite{peters2015solving,gupta2015phasor,heide2013low,kadambi2013coded}. Interferometric approaches have demonstrated micron-scale temporal light transport~\cite{gkioulekas2015micron}.
Combining these two threads of spatial and temporal dimensions makes it possible to reason over rich decompositions that encode how light travels at a given time and position.
Such joint spatio-temporal analysis has been implemented using frequency-aware probing methods with a correlation sensor-illumination pair~\cite{o2014temporal} or, recently, using an interferometric setup with programmable amplitude and phase modulators~\cite{kotwal2020interferometric}.
The spatio-temporal transport structure enables direct-indirect separation for multipath interference removal~\cite{naik2015light,whyte2015resolving,agresti2018combination}.
Also, existing works have explored joint analysis of temporal and partial polarimetric dimensions for light field imaging~\cite{manakov2013reconfigurable}, surface reconstruction~\cite{maeda2018dynamic,dashpute2018depth} and imaging through scattering {media}~\cite{wu2016scattering,wu2018adaptive}.

None of these existing methods are capable of capturing full polarization states (linear, circular, ellipsoidal with varying degrees) jointly with spatial and temporal dimensions. The only polarimetric sampling demonstrated before captures partial linear polarization~\cite{gkioulekas2015micron}. In this work, we propose a novel spatially- and temporally-resolved ellipsometry methods that allow us to capture a complete polarimetric light transport of any polarization state. As such, for the first time, we demonstrate full spatio-polarimetric and temporal-polarimetric captures of the 7-dimensional light transport (2D spatial, 2D detector, 2D polarimetric, 1D temporal dimension).

\paragraph{Polarimetric Imaging in Graphics and Vision}
The polarization state of light has been exploited by a broad range of methods in computer graphics and vision to encode material-dependent scene properties or source-dependent transport.
Cross-polarization imaging, one of the most common polarimetric acquisition approaches, uses either linear or circular polarization filters in front of a camera and a light source with perpendicular orientation which separates captured light into two components depending on whether light maintains the polarization of the light source or not.
This approach to source separation has been adopted in various applications, most notably seeing through scattering {media}~\cite{treibitz2008active,gupta2008controlling,tanaka2013descattering} and its application for shape estimation~\cite{chen2007polarization}, also diffuse-specular separation~\cite{ma2007rapid,ghosh2008practical,ghosh2010circularly,nagano2015skin,riviere2017polarwild} for facial capture and geometric scene understanding.
This relies on unique polarization encoding of material-dependent reflectance properties.
Polarimetric imaging also has been used for estimating depth using birefringent materials~\cite{baek2016birefractive,meuleman2020single} and surface normals~\cite{atkinson2006recovery,kadambi2015polar,baek2018simultaneous,deschaintre2021deep,ichikawa2021shape} in addition to metallic surface characterization~\cite{berger2011modeling}.
While substantial progress has been made in exploiting polarization for these diverse applications, the information of \emph{all} polarization states jointly, along with spatial and temporal resolutions, has not been exploited in graphics and vision. {Indeed, only very recently researchers proposed data-driven polarimetric BRDF models~\cite{baek2020image}, however, without considering the spatial and temporal dimensions}.
In this work, we exploit rich structure in light polarization along with space and time dimensions, and propose a complete spatio- and temporal-polarimetric imaging and decomposition method that allows us to analyze scene properties hidden to existing light transport imaging methods.

\paragraph{Optical Ellipsometry}
Consider a material sample observed by a sensor and illuminated by a light source. The polarimetric response of the sample encodes material information including its refractive index, roughness, composition, crystalline structure, and electrical conductivity.
Ellipsometry describes a method of obtaining such material properties by capturing the complete polarimetric transport that the material sample is subject to, often by synchronously modulating the polarization states of emitted and captured light. Ellipsometric methods have been studied during the past few decades in optics~\cite{azzam2016stokes}, material science~\cite{jellison2018crystallographic,ramsey1994influences}, and biology~\cite{ghosh2011tissue,arwin2011application}.
In this work, we go beyond conventional ellipsometry by jointly capturing spatial, temporal, and polarimetric dimensions.
As a result, the proposed spatio- and temporal-polarimetric imaging allows us to decompose light transport into unprecedented detail, revealing joint polarimetric spatial and temporal information that appears ``hidden'' to existing methods.
In addition, we analyze low-rank structure of real-world polarimetric light transport and develop novel data-driven ellispometry methods which rotate polarizing optics at optimal angles to achieve high-fidelity acquisition of polarimetric light transport with {a} reduced number of captures.

\section{Background}
\label{sec:background}
Before introducing the proposed spatio- and temporal-polarimetric light transport imaging method, we review existing models for spatial and temporal light transport in graphics and vision, and polarimetric light transport in optics.

\paragraph{Spatial Light Transport}
Given a 2D projector and a 2D camera, the spatial light transport between them can be modeled as a matrix-vector multiplication~\cite{ngan2005,sen2005dual}
\begin{equation}\label{eq:spatial_light_transport}
  \mathbf{i}_s = \mathbf{T}_s\mathbf{p}_s,
\end{equation}
where $\mathbf{i}_s, \mathbf{p}_s \in \mathbb{R}^{S \times 1}$ are the vectorized 2D images captured and emitted by the camera and the projector, respectively, and $S$ being the number of pixels. The matrix $\mathbf{T}_s \in \mathbb{R}^{S \times S}$ is the spatial light transport matrix that describes how light intensity, emitted at a specific angle from a projector pixel, changes as it travels through a scene and is captured at a specific pixel on the camera. The matrix $\mathbf{T}_s$, therefore, decomposes the \emph{spatial dimension} of light transport indexed as pixels on the {projector} and camera, respectively, see Figure~\ref{fig:LT_matrix}(a).

\paragraph{Temporal Light Transport}
Spatial light transport assumes a steady-state equilibrium of global light transport. However, in reality, emitted light {reaches the detector along potentially complex paths with varying lengths, and hence may arrive at the sensor at different travel times}. This temporal light transport, before reaching a steady state, can be modeled for a \emph{single detector pixel} as
\begin{equation}\label{eq:temporal_light_transport}
\begin{array}{l}
\mathbf{i}_t = \mathbf{T}_t p^{t},
\end{array}
\end{equation}
where $\mathbf{i}_t \in \mathbb{R}^{\Gamma \times 1}$ is the vector of detected intensities of travel time $t$ discretized in $\Gamma$ different temporal bins.
The scalar $p^{t}$ describes the emitted light intensity at the time index zero. The matrix $\mathbf{T}_t \in \mathbb{R}^{\Gamma \times 1}$ is \emph{temporal light transport matrix} for a given detector pixel, see Figure~\ref{fig:LT_matrix}(b).
This can be further generalized to {shift}-invariant light transport with the convolutional model
\begin{equation}\label{eq:temporal_light_transport_conv}
\begin{array}{l}
\mathbf{i}_t =\mathbf{T}_t * \mathbf{p}_{t},
\end{array}
\end{equation}
where $*$ is a matrix-vector convolution~\cite{o2014temporal} across the temporal dimension $t$ and {$\mathbf{p}_t \in \mathbb{R}^{\Gamma \times 1}$ are the intensities vectorized in time bins.}

\begin{wrapfigure}{r}{0.4\linewidth}
  \vspace{0mm}
  \centering
  \includegraphics[width=\linewidth]{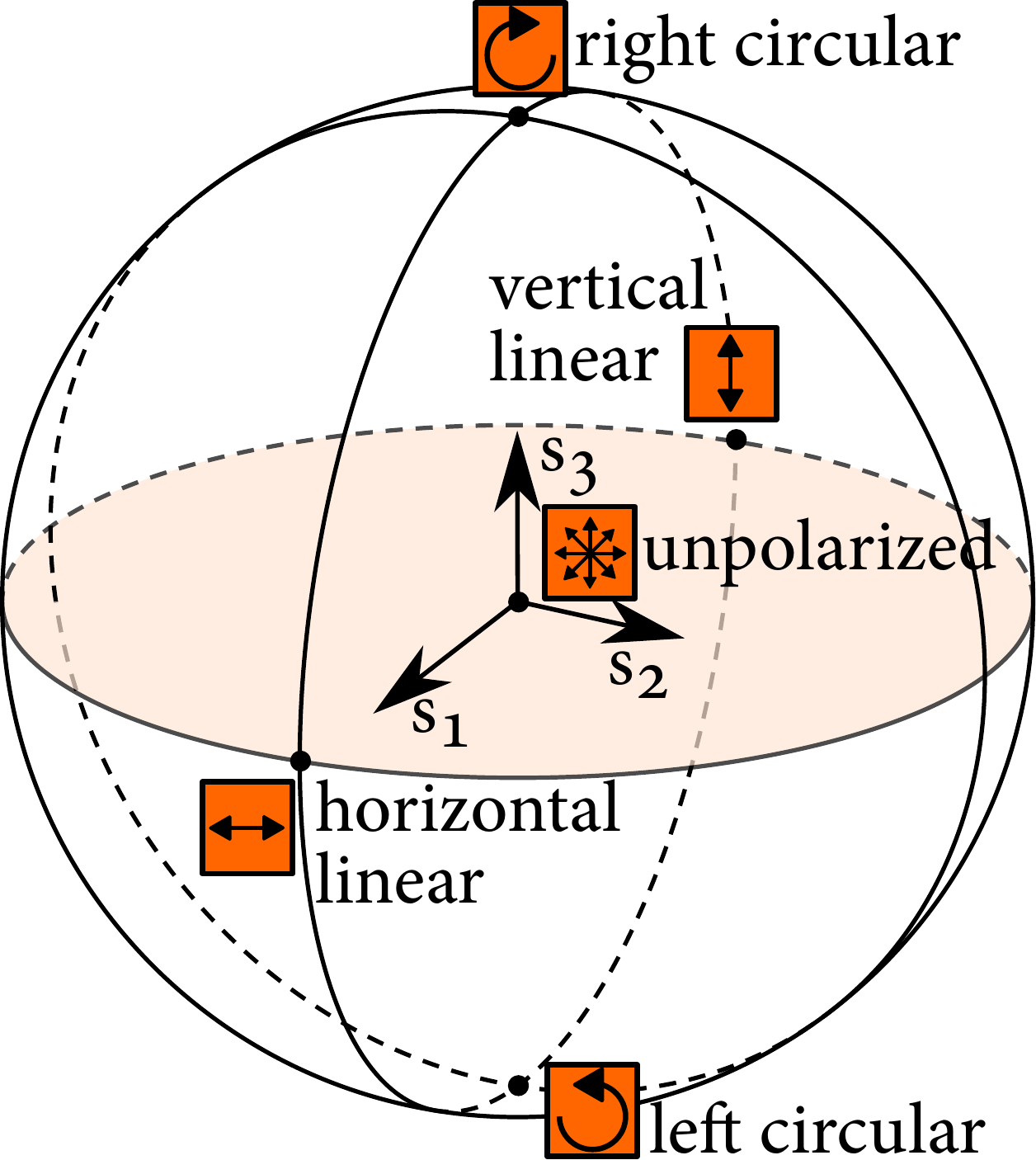}
  \caption{\label{fig:poincare}%
  {Poincar\'{e}} sphere.
  \vspace{-3mm}
  }
\end{wrapfigure}

\paragraph{Polarimetric Light Transport}
Polarization is an intrinsic wave property of light describing the geometric oscillation pattern of the electromagnetic field, i.e., light, in time. The Stokes-Mueller formalism provides a representation for polarization, describing any polarization state of light and its change~\cite{collett2005field}. Specifically, the Stokes vector is a four-dimensional vector describing the polarization state of light: $\mathbf{s} = [s_0, s_1, s_2, s_3]$, where each element describes {the} radiance of light, {the} linear-polarization components at $0^\circ$ and $45^\circ$, and {left-right} circular-polarization component.

To provide intuition for the Stokes vector formalism, Poincar\'{e} spheres offer an illustrative visualization of the complete polarization state of light based on the Stokes vector by plotting the last three Stokes elements normalized by the total power $s_0$: $[s_1/s_0, s_2/s_0, s_3/s_0]$, see Figure~\ref{fig:poincare}. A Stokes vector is then represented as a vector inside the Poincar\'{e} sphere. {The sphere's center represents unpolarized light and light becomes more polarized as we are approaching the surface of the Poincar\'{e} sphere.} The direction of the Stokes vector in the Poincar\'{e} sphere describes the type of polarization such as horizontal linear or right circular polarization.

To characterize polarimetric transport, {we use the Mueller matrix which describes the polarimetric response of a scene as the change of a Stokes vector}
\begin{equation}\label{eq:polarimetric_light_transport}
  \mathbf{i}_p = \mathbf{T}_p\mathbf{p}_p,
\end{equation}
where $\mathbf{i}_p$, $\mathbf{p}_p$ are the Stokes vectors of the detected and emitted light, and $\mathbf{T}_p$ is the polarimetric light transport matrix in the form of a Mueller matrix for a single detector pixel, see Figure~\ref{fig:LT_matrix}(c).

\section{Tensor Light Transport}
Existing spatial~\cite{o2012primal}, temporal~\cite{velten2013femto}, and polarimetric~\cite{collett2005field} light transport models rely on matrix-vector representations.
While these linear models offer convenient links to linear algebra for each plenoptic dimension, they do not treat {light transport jointly} across space, time, and polarization. As a generalization to these existing models, we \emph{propose a tensor-based model of light transport}.
For a primer on tensor algebra, we refer the reader to Itskov~\shortcite{itskov2007tensor}.

We start by modeling the polarization state and intensity of emitted pulsed light as {a} 3D tensor $\mathcal{P}(s,p)$, parameterized by the 2D spatial direction $s$ and the 1D Stokes vector dimension $p$. Light modeled {by} the tensor $\mathcal{P}$ then travels through a scene, e.g., undergoing potentially complex scattering effects, and encode{s} material and geometric properties of the scene. When light returns to the detector, we model it as a 4D tensor $\mathcal{I}(s,p,t)$.
The temporal dimension $t$ (1D) denotes the travel time, $s$ is the camera pixel location (2D) and $p$ is the Stokes vector dimension (1D).
With the two tensors $\mathcal{P}$ and $\mathcal{I}$ in hand, we model the spatio-temporal polarimetric light transport as a 7D tensor $\mathcal{T}\left(s,s',p,p',t\right)$, where $s$ and $s'$ are the 2D spatial sample locations at the detector and the light source, $p$ and $p'$ are the 1D Stokes vectors of the detected and the emitted light, and $t$ is the 1D travel time.
Refer to Figure~\ref{fig:LT_matrix}(d).
We write the light transport equation using tensor contraction~\cite{itskov2007tensor} as
\begin{equation}\label{eq:general_light_transport}
  \mathcal{I}\left(s,p,t\right) = \sum\limits_{s',p'} \mathcal{T}\left(s,s',p,p',t\right)\mathcal{P}\left(s',p'\right).
\end{equation}
For {shift}-invariant light transport, this can be formulated as a convolutional integral
\begin{equation}\label{eq:general_light_transport}
  \mathcal{I}\left(s,p,t\right) =  \sum\limits_{s',p'} \int_{-\infty}^{\infty} \mathcal{T}\left(s,s',p,p',t-t'\right)\mathcal{P}\left(s',p',t'\right)dt',
\end{equation}
where $t'$ is the time dimension of the {time-resolved} illumination tensor $\mathcal{P}$.

\begin{figure}[t]
  \centering
  \includegraphics[width=0.95\linewidth]{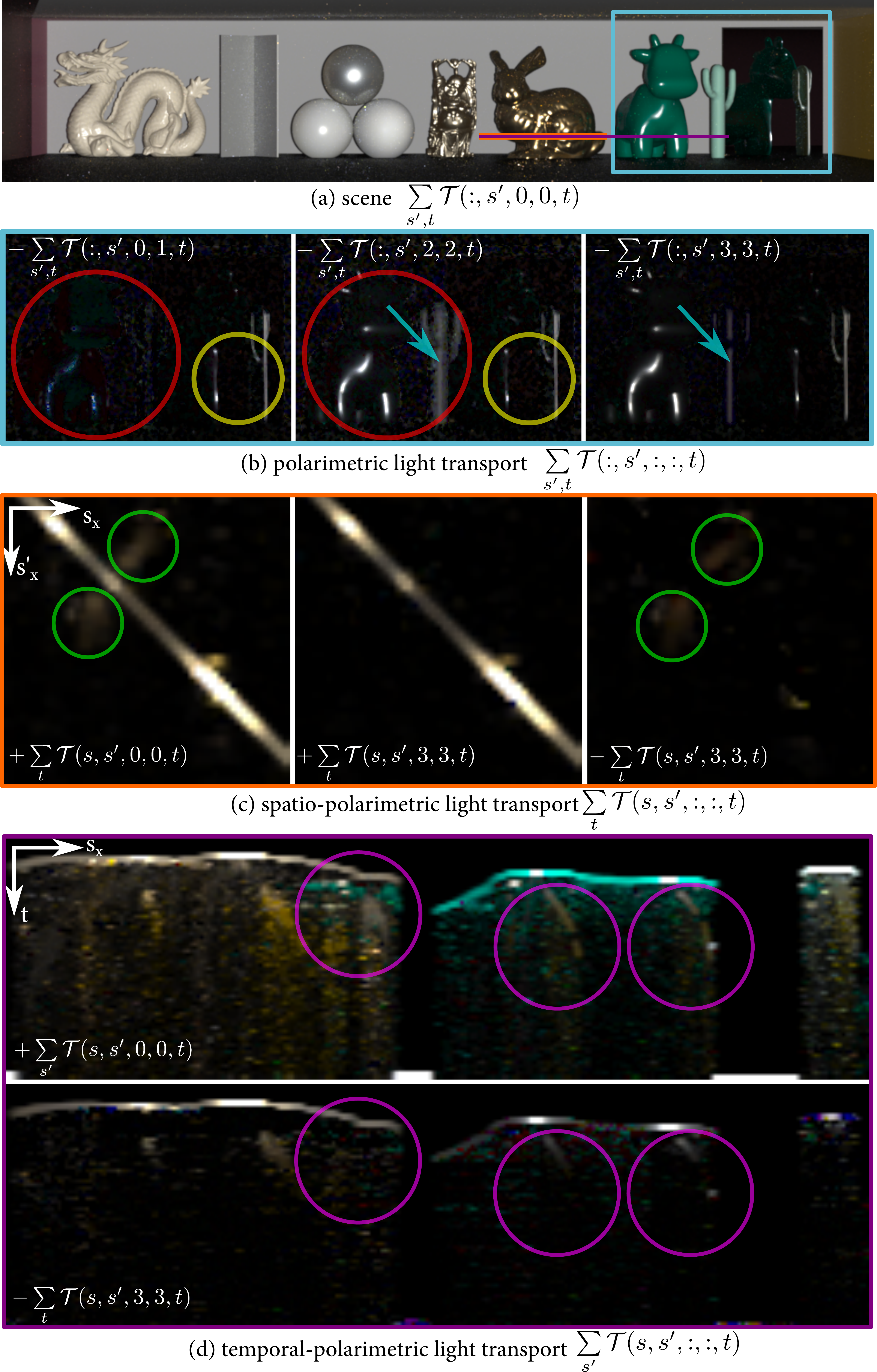}
  \vspace{-3mm}
  \caption{\label{fig:LT_render}%
	Rendered simulation of a complete 7D light transport $\mathcal{T}$ tensor using a multi-dimensional path tracer.
  (a) A conventional camera integrates light intensity emitted from all projector pixels $s'$ at temporal steady-state $t$, capturing $\sum _{s',t} {{\mathcal{T}}(:,s',0,0,t)}$.
  We analyze the tensor $\mathcal{T}$ by slicing different dimensions of space, time, and polarization.
  (b) The polarimetric light transport $\sum _{s',t} {{\mathcal{T}}(:,s',:,:,t)}$ reveals material-dependent scene responses, for instance, untangling reflections from diffuse surface points (red circles) and mirrors (yellow circles), and allows us to separate single-bounce specular reflection from scattered specular reflection (cyan arrows).
  (c) An additional spatial dimension, spatio-polarimetric light transport $\sum _{t} {{\mathcal{T}}(s,s',:,:,t)}$ separates single-bounce specular reflection from multi-bounce specular reflection (green circles) visualized for the orange scanline shown in (a).
  (d) Capturing the temporal-polarimetric light transport $\sum _{s'} {{\mathcal{T}}(s,s',:,:,t)}$ makes it possible to {decompose} time-varying global illumination into diffuse and specular components (magenta circles) visualized for the purple scanline shown in (a).
  \vspace{-5mm}
  }
\end{figure}
This 7D spatio-temporal polarimetric light transport tensor $\mathcal{T}$ describes the complete polarimetric change of light for every combination of camera and projector pixels with high temporal resolution of the travel time.
{\textit{Probing} operations can be performed by slicing the tensor along one or two indices, reducing the tensor into a vector or a matrix.
This means that our tensor-based transport model is backward-compatible to the existing matrix representations of light transport in Equations~\eqref{eq:spatial_light_transport}, \eqref{eq:temporal_light_transport}, \eqref{eq:polarimetric_light_transport} as follows}
\vspace{-8mm}
\begin{align}\label{eq:tensor2pol}
  \mathbf{T}_s &= \sum_{t}\mathcal{T}(:,:,0,0,t), \nonumber \\
  \mathbf{T}_t &= \sum_{s,s'}\mathcal{T}(s,s',0,0,:), \nonumber \\
  \mathbf{T}_p &= \sum_{s,s',t}\mathcal{T}(s,s',:,:,t).
\end{align}

\subsection{Path Tracer for 7D Light Transport Simulation}
To provide further intuition on the properties of the high-dimensional 7D light transport tensor $\mathcal{T}$, we developed a path tracer and {we} simulate a synthetic capture system consisting of a high-speed 2D projector{,} a high-speed 2D camera with picosecond temporal resolution, and polarizing optics. We note, that although such high fidelity transient projector{s} and sensor{s} do not exist today.
{We validate the proposed method using} two practical imaging methods in the remainder of this work.
We implement our renderer on top of the Mitsuba2 renderer~\cite{nimier2019mitsuba}.
Specifically, we set up spatial acquisition through masking operations for a projector-camera pair.
For the temporal dimension, we keep track of the time-of-flight for each light ray throughout the rendering process~\cite{jarabo2014framework,pediredla2019ellipsoidal}.
For polarization, we use a real-world polarimetric BRDF dataset~\cite{baek2020image} and implement a dual-rotating-retarder setup~\cite{azzam1978photopolarimetric}.
We render trichromatic imagery at wavelengths of $450$, $550$, and $660$\,nm.
Figure~\ref{fig:LT_render} shows the rendered 7D light transport, revealing geometry- and material-dependent reflections. We observe that by separating polarization, space, and time dimensions, it is possible to distinguish mirror and scene reflections, acquire mirror-like components isolated from cluttered specular reflection, decompose specular reflections into single and multi-bounces, and analyze global illumination as mixed diffuse and specular transport.

\begin{figure}[t]
  \centering
  \includegraphics[width=\linewidth]{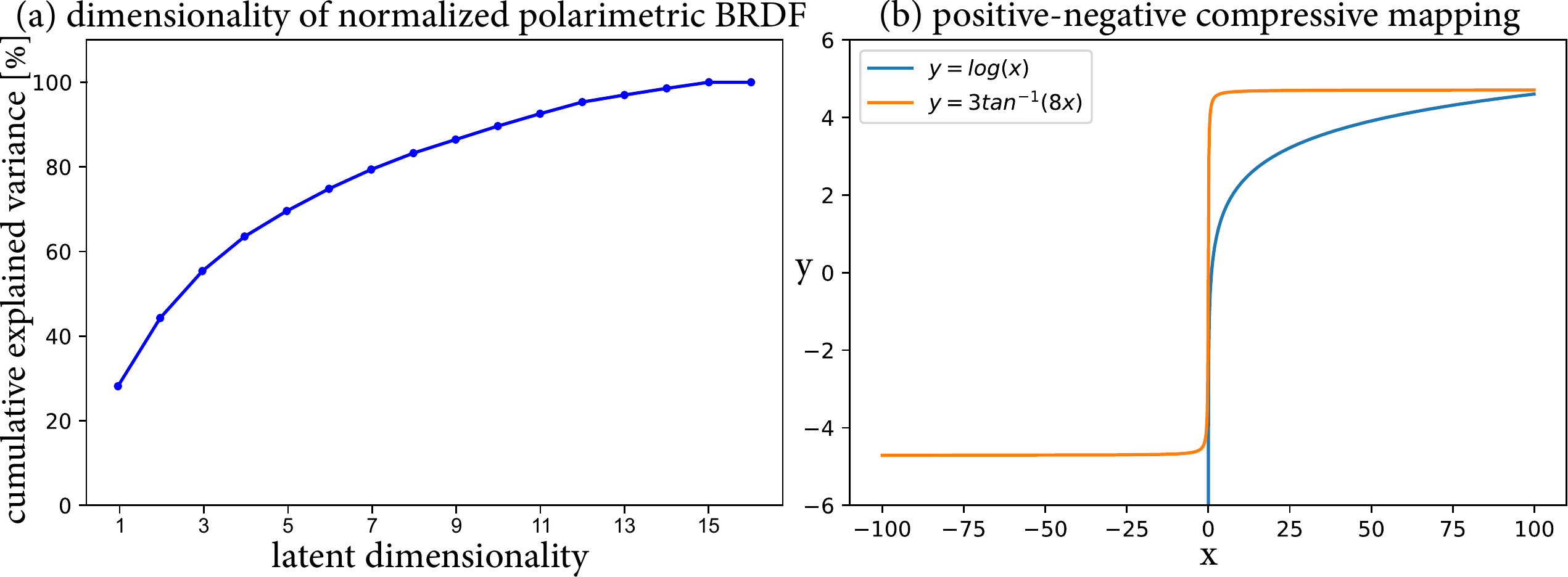}
  \caption{\label{fig:pca_pbrdf}%
  {PCA analysis of polarimetric light transport via real-world materials. (a) Polarimetric BRDFs with intensity normalization exhibit low-rank structure. (b) To perform PCA analysis, we propose the arctangent mapping that handles both negative and positive values of polarimetric light transport, in contrast to conventional positive-only log mappings.}
  }
  \vspace{-5mm}
\end{figure}

\subsection{Low-dimensional {Polarimetric Transport} Embeddings}
We further analyze the light transport tensor, in particular its polarization slice, which has been overlooked in graphics and vision compared to the other well-understood dimensions, such as space and time.
Specifically, we analyze low-dimensional polarimetric embeddings of the high-dimensional light transport tensor $\mathcal{T}$. Such lower-dimensional embeddings offer statistical priors that may allow {for} efficient acquisition and provide insight into the transport properties.
Low-dimensional transport structure has {also} been {used} in the spatial dimension~\cite{o2010optical}, spectral dimension~\cite{saragadam2019krism}, and temporal dimension~\cite{heide2019non,o2018confocal}.

To this end, we perform {principal} component analysis (PCA) on 25 real-world polarimetric BRDFs~\cite{baek2020image}.
{
This facilitates efficient polarimetric analysis, in contrast to na\"ively analyzing a large collection of rendered polarimetric images using costly polarization-aware path tracing.
Furthermore, polarimetric light transport can be represented as an intensity-scaled polarimetric BRDF by integrating over indirect transport components.
}

Inspired by PCA analysis on intensity BRDFs~\cite{nielsen2015optimal}, we develop our analysis method with two major distinctions.
First, we normalize the polarimetric BRDF in order to decouple polarization from intensity via division by the intensity component: $\mathbf{T}_p \mapsto \mathbf{T}_p/\mathbf{T}_p(0,0)$. This {allows} us to {separately} investigate the polarimetric variation {independently} of intensity scale.
Second, to distribute the polarimetric energy across {a} high dynamic range of inputs, we propose to apply the arctangent mapping: $\mathbf{T}_p \mapsto \arctan(c\mathbf{T}_p)$, where $c$ is a scalar which is set to 8 in our experiment. This handles positive and negative inputs in contrast to the positive-only log mapping~\cite{nielsen2015optimal}.
We arrange the normalized and compressed polarimetric BRDFs to form an observation matrix {$\mathbf{X}\in\mathbb{R}^{N\times 16}$, where $N$ is the number of valid angular bins for all dataset materials.}
We extract the principal components of $\mathbf{X}$ using singular value decomposition (SVD)
\begin{equation}\label{eq:pbrdf_SVD}
  \mathbf{X}-\mathbf{\mu} = \mathbf{U}\mathbf{\Sigma}\mathbf{V}^\intercal,
\end{equation}
where $\mathbf{\mu}$ is the mean matrix of $\mathbf{X}$ over {the valid angular bins} of different materials. $\mathbf{U}$ and $\mathbf{V}$ contain the eigenvectors and $\mathbf{\Sigma}$ has the eigenvalues.
Then, we obtain the scaled principal components $\mathbf{V}\mathbf{\Sigma}$ shown in Figure~\ref{fig:pca_pbrdf}. {These visualizations reveal the low-rank structure in the captured light transport.
We refer to the Supplemental Document for a low-rank analysis of the angular-polarimetric dimensions.}

\section{Learning Rotating Ellipsometry}
\label{sec:learned_ellipsometry}
To acquire the complete polarimetric light transport, conventional optical ellipsometry rotates polarizing optics in front of a light source and a sensor while {acquiring} polarization-modulated intensities from which polarimetric light transport is reconstructed.
For instance, dual-rotating-retarders (DRR)~\cite{azzam1978photopolarimetric} uses two fixed linear polarizers and two rotating retarders at hand-crafted angles, resulting in many {captures}, typically 36.
The previous section illustrates low-dimensional polarimetric embeddings in the transport tensor which opens a potential avenue to a more efficient capture procedure.
A similar principle has been applied to the plenoptic dimensions defined on intensity by iteratively acquiring low-rank approximations of respective light transport slices~\cite{o2010optical,saragadam2019krism}.
However, \emph{directly capturing low-dimensional polarimetric structure is infeasible} as polarimetric light transport is based on Stokes-Mueller formalism, not intensity measurements, therefore an additional ellipsometric reconstruction step is required.
To overcome this challenge, we propose a data-driven ellipsometry method {that} implicitly exploits this polarimetric low-rank structure for efficient capture.
We dub our method {\emph{learned quad-rotating polarizer retarder}} (L-QRPR).
It uses rotating linear polarizers and retarders at learned angles to provide accurate reconstructions for real-world polarimetric light transport tensors.

\begin{figure}[t]
  \centering
  \includegraphics[width=\linewidth]{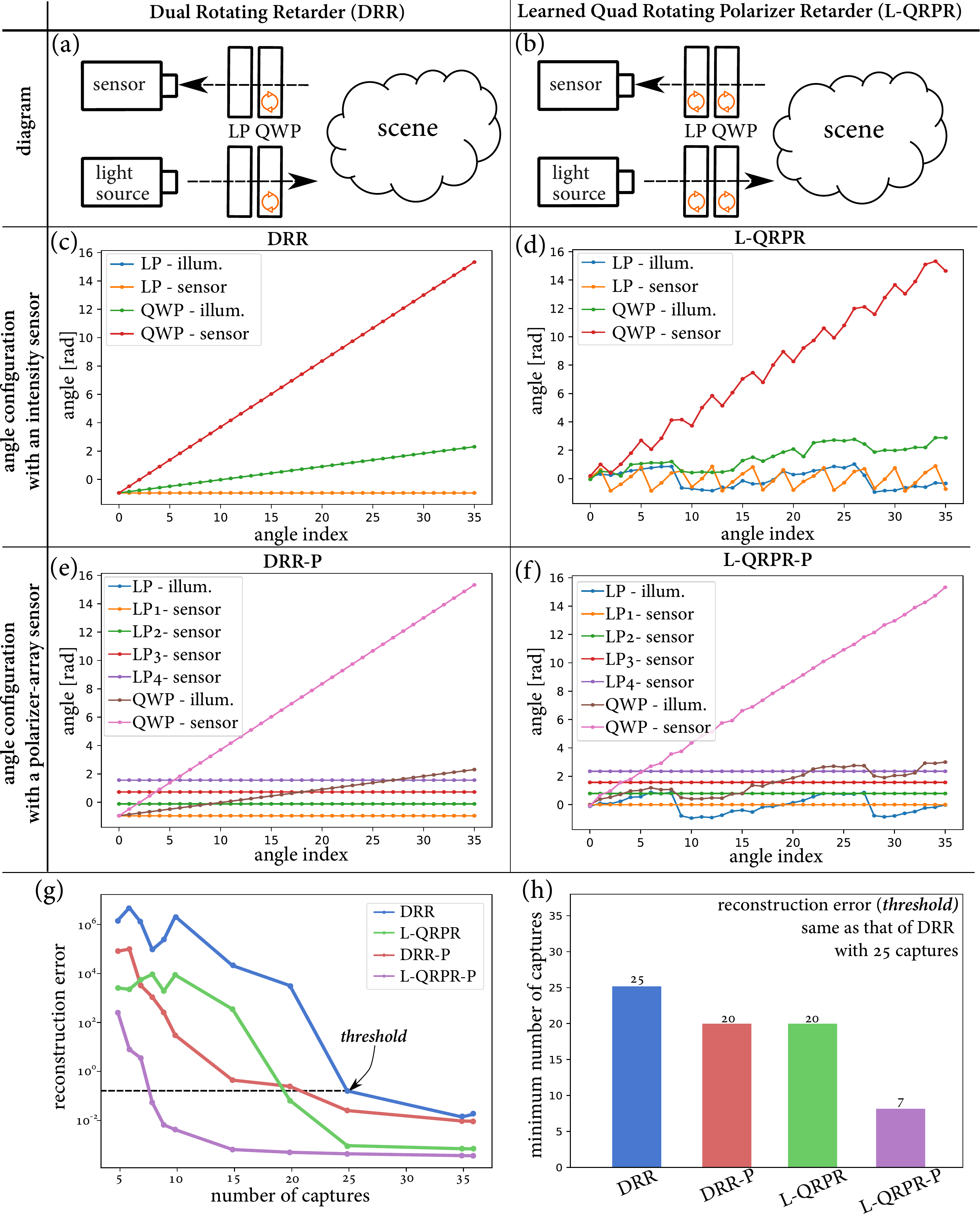}
  \caption{\label{fig:ellipsometry_opt}%
  (a) Conventional DRR ellipsometry rotates the two quarter-wave plates at (c) manually-chosen angle sets.
  (b) In contrast, we propose to rotate the two linear polarizers and the two retarders (d) using a smaller set of angles that learns to exploit the low-rank structure of natural polarimetric light transport.
  (f) We also show how the proposed method can be extended to an emerging polarizer-array sensor.
  {To evaluate the proposed learned ellipsometry method, we conduct 5-fold cross-validation.}
  (g) Our learned ellipsometry consistently outperforms the conventional DRR method both for the intensity sensor and the polarizer-array sensor.
  (h) Our method substantially reduces the number of required captures by more than $3\times$ without scarifying reconstruction accuracy.
  \vspace{-7mm}
  }
\end{figure}
To learn the set of optimal angles, we first propose a differentiable image formation model of a rotating ellipsometry
\begin{equation}\label{eq:rotating_ellipsometry}
I^{\theta_k^4, \theta_k^3, \theta_k^2, \theta_k^1} = \left[\mathbf{L}^{\theta_k^4}\mathbf{Q}^{\theta_k^3}\mathbf{T}_{p}\mathbf{Q}^{\theta_k^2}\mathbf{L}^{\theta_k^1}\mathbf{p}\right]_0 + \eta,
\end{equation}
where $k$ is the rotation index out of the total number of captures $K$. $\mathbf{p}=[1,0,0,0]^\intercal$ is the unit Stokes vector and $\left[\cdot\right]_0$ is the first element of the vector.
$\mathbf{Q}^{\theta}$ and $\mathbf{L}^{\theta}$ are the Mueller matrices of a quarter-wave plate (QWP) and a linear polarizer (LP) at an angle $\theta$.
For the definitions of the Mueller matrices, we refer to the Supplemental Document.
$I^{\theta_k^4, \theta_k^3, \theta_k^2, \theta_k^1}$ is the simulated intensity and $\eta$ is Gaussian noise with standard deviation $5\times10^{-4}$.
Note that the matrix-vector multiplications based on Stokes-Muller formalism make this image formation naturally differentiable with respect to the angles of the polarizing optics, $\theta_{k\in\{1,\cdots,K\}}^{i\in\{1,2,3,4\}}$.

Given the simulated intensities, we reconstruct the polarimetric light transport by solving an optimization problem
\begin{equation}\label{eq:recon_mueller}
\mathop {\mathrm{minimize} }\limits_{\hat{\mathbf{T}}_{p}} \sum\limits_{k = 1}^{K} {\left( I^{\theta_k^4, \theta_k^3, \theta_k^2, \theta_k^1} - \left[\mathbf{L}^{\theta_k^4}\mathbf{Q}^{\theta_k^3}\hat{\mathbf{T}}_{p}\mathbf{Q}^{\theta_k^2}\mathbf{L}^{\theta_k^1}\mathbf{p}\right]_0\right)}^2,
\end{equation}
where $\hat{\mathbf{T}}_{p}$ is the estimate of the polarimetric light transport.
Note that the solution to Equation~\eqref{eq:recon_mueller} can be analytically obtained {as} the least-square solution~\cite{baek2020image}, and is therefore differentiable.

Last, we compute the loss with the ground truth $\mathbf{T}_p$, and optimize the angle set to minimize this loss
{
\begin{equation}\label{eq:angle_opt}
  \mathop{\mathrm{minimize}}\limits_{\Theta_k^4, \Theta_k^3, \Theta_k^2, \Theta_k^1} \mathcal{L}(\hat{\mathbf{T}}_p, \mathbf{T}_p),
\end{equation}
}
where $\mathcal{L}$ is the mean-squared-error loss function {and $\Theta_k^i=\{\theta_1^i, ..., \theta_K^i\}$ is an angle set for each optical element}.
We use the Adam optimizer~\cite{paszke2017automatic} and initialize the angles of polarizing optics following the conventional DRR setting~\cite{azzam1978photopolarimetric} as  $\theta_k^1=0^{\circ},\theta_k^2=5(k-1)^{\circ},\theta_k^3=25(k-1)^{\circ},\theta_k^4=0^{\circ}$ with $k\in\{1,2,\cdots,K\}$.
As ground truth data, we use the real-world polarimetric BRDF dataset from {Baek et al.~\shortcite{baek2020image}}.
{
Note that data-driven optimal capture has also been studied for reflectance acquisition~\cite{nielsen2015optimal,xu2016minimal}.
}

\paragraph{Learning Rotation Angles with a Polarizer-array Sensor}
The proposed method {allows for} learning optimal angles not only for the conventional CMOS intensity sensor{s}, but also for emerging polarizer-array CMOS sensor{s}, e.g., Sony IEDM2016.
{These emerging sensors} increase {the} sampling rate by four times {by using} on-sensor linear polarizers with four different orientations ($0^\circ, 45^\circ, 90^\circ, 135^\circ$).
We incorporate these hardcoded polarization directions into the differentiable image formation as
\begin{equation}\label{eq:rotating_ellipsometry}
I_{q\in{\{1,2,3,4\}}}^{\theta_k^4, \theta_k^3, \theta_k^2, \theta_k^1} = \left[\mathbf{L}_{q\in{\{1,2,3,4\}}}\mathbf{Q}^{\theta_k^3}\mathbf{T}_{p}\mathbf{Q}^{\theta_k^2}\mathbf{L}^{\theta_k^1}\mathbf{p}\right]_0 + \eta,
\end{equation}
where $q$ is the index of the sensor linear polarizers.
Inserting this {into} Equation~\eqref{eq:recon_mueller} and Equation~\eqref{eq:angle_opt} allows us to optimize for the angle sets of $\theta_k^3, \theta_k^2, \theta_k^1$ while the polarizer angles for the sensor are fixed.

Figure~\ref{fig:ellipsometry_opt} shows that our ellipsometry methods, {learned quad-rotating polarizer and retarder (L-QRPR) and its variant with a polarization camera (L-QRPR-P), outperform the conventional dual rotating retarder (DRR)} method in terms of capture efficiency.
{For synthetic validation, we conduct a 5-fold cross-validation experiment.} {L-QRPR and L-QRPR-P both achieve the same accuracy as conventional DRR using 25 captures while using 20\% and 72\% fewer captures respectively.}
As such, we validate that the proposed data-driven rotating ellipsometry effectively exploits the low-rank structure in natural polarimetric light transport.
{
For experimental validation, we use the angles learned from the pBRDF dataset~\cite{baek2020image} and show that our method generalizes to unseen materials, see Figures~\ref{fig:learned_ellipsometry_eval} and \ref{fig:learned_ellipsometry_scene}.
}
We refer to the Supplemental Document for additional assessment.

\begin{figure}[t]
  \centering
  \includegraphics[width=\linewidth]{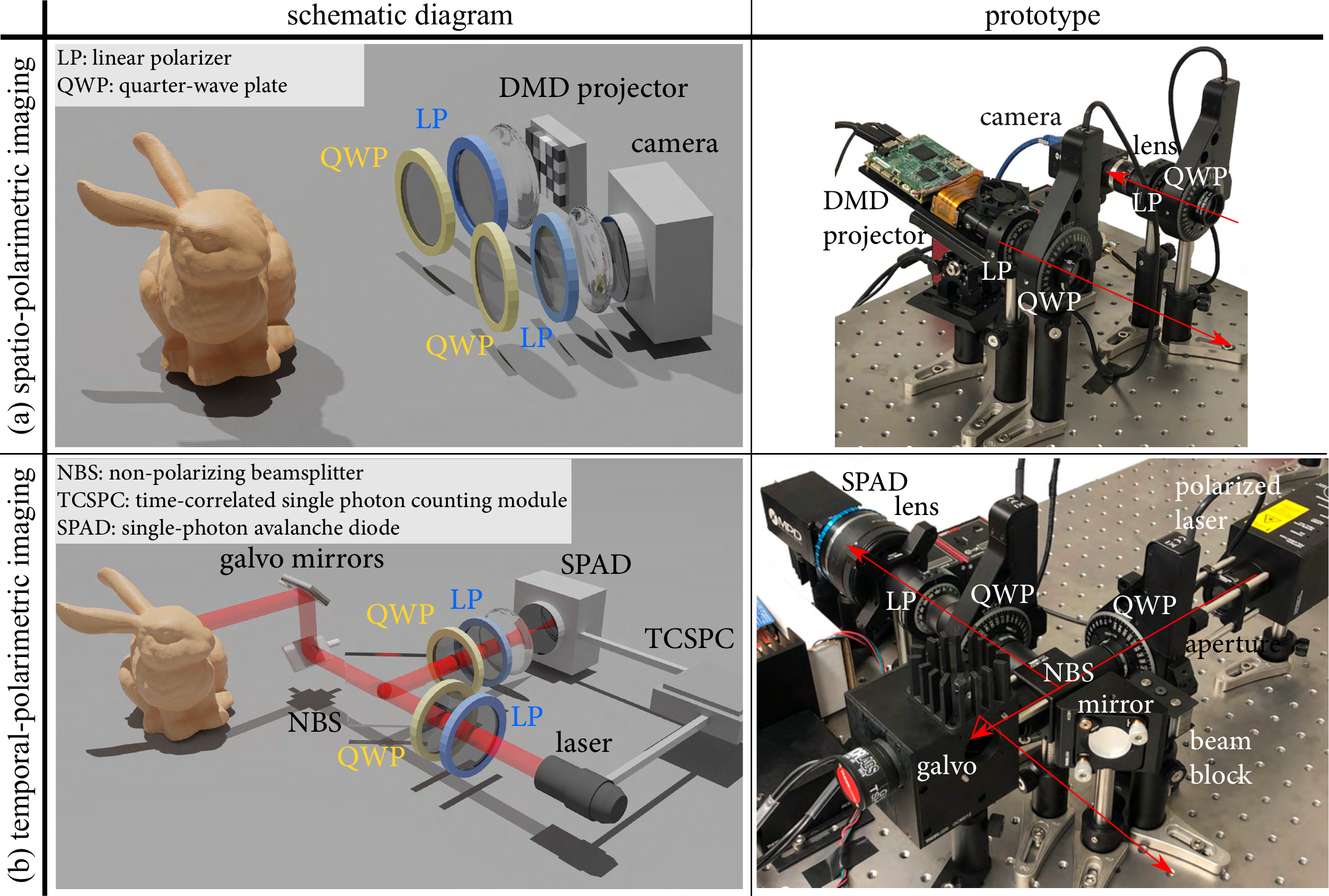}
  \caption{\label{fig:imaging}%
  We propose two novel imaging systems that capture spatio-polarimetric light transport $\mathcal{T}_{sp}$ and coaxial temporal-polarimetric light transport $\mathcal{T}_{pt}$.
  The spatio-polarimetric system acquires the complete polarimetric scene response for every combination of camera and projector pixels over trichromatic spectral bands, using camera-projector masks with rotating polarization optics.
  The temporal-polarimetric setup captures the coaxial spatial dimension at picosecond temporal resolution for a narrow spectral band, also with the full polarimetric resolution.
  We scan the transport using {a} 2-axis galvo mirror and rotating QWPs.
  }
  \vspace{-5mm}
\end{figure}

\section{Spatio-Temporal Polarimetric Imaging}
\label{sec:ellipsometric_spatio_temporal}
The full 7D light transport tensor $\mathcal{T}$ is challenging to acquire due to its high dimensionality of space, time, and polarization, even with our learned ellipsometry.
While methods have been proposed before that acquire transport for each individual dimension, the joint acquisition and analysis of {the full} 7D light transport has been a challenge, especially for the polarization dimension.
We propose polarimetric acquisition approaches that allow us to capture space, time and polarization dimensions of light transport jointly, while being realizable as practical imaging systems.
{The absence of high-resolution, time-tagged sensors and pulsed projectors poses a practical restriction on capture time.}
To address this challenge, we propose two novel computational imaging systems that capture spatio-polarimetric light transport $\mathcal{T}_{sp}$ (6D) and coaxial temporal-polarimetric light transport $\mathcal{T}_{pt}$ (5D) respectively
\begin{align}\label{eq:tensor2pol}
  \mathcal{T}_{sp} &= \sum \limits_{t}\mathcal{T}(:,:,:,:,t), \nonumber\\
  \mathcal{T}_{pt} &= \mathcal{T}(s,s,:,:,:), \forall s.
\end{align}

In the following, we describe these two instantiations of our generalized polarimetric transport acquisition.
We refer the reader to the Supplemental Document for comparison with other systems and calibration details.

\subsection{Spatio-Polarimetric Imaging}
We present a computational imaging method that captures the 6D spatio-polarimetric light transport $\mathcal{T}_{sp}$ using a high-resolution intensity sensor.
The 2D intensity sensor allows us to efficiently resolve spatial dimensions of the 2D illumination $s'$ and camera $s$.
Figure~\ref{fig:imaging}(a) illustrates this spatially-multiplexed acquisition system.

\paragraph{Illumination}
To spatially multiplex illumination patterns, we employ a DMD projector that emits unpolarized light with a programmable intensity from a certain projector pixel $s'$.
Light polarization is then modulated by passing through a LP and a QWP.
The LP and the QWP are mounted on rotation stages at angles $\theta_k^1$ and $\theta_k^2$, respectively at the $k$-th capture index.
This provides the emitted light tensor $\mathcal{P}^{\theta_k^2,\theta_k^1}({s'},:)$ from a projector pixel $s'$
\begin{equation}\label{eq:illumination_stokes_transient}
\mathcal{P}^{\theta_k^2,\theta_k^1}({s'},:) = \mathbf{Q}^{\theta_k^2}\mathbf{L}^{\theta_k^1}\mathbf{p}^{s'},
\end{equation}
where $\mathbf{p}^{s'}=[p^{s'},0,0,0]^\intercal$ is the Stokes vector of the unpolarized DMD light and $p^{s'}$ is the intensity for that spatial direction $s'$.

\paragraph{Spatio-Polarimetric Light Transport}
The emitted light interacts with a scene described by the spatio-polarimetric light transport $\mathcal{T}_{sp}$ that is
$\mathcal{R}^{\theta_k^2,\theta_k^1}(s,:) = \mathcal{T}_{sp}(s,s',:,:)\mathcal{P}^{\theta_k^2,\theta_k^1}(s',:)$,
where $s$ is the spatial position on the detection module.
The spatio-polarimetric light transport $\mathcal{T}_{sp}$ assumes steady-state equilibrium of light transport, integrating the full tensor $\mathcal{T}$ over the temporal dimension $t$.

\paragraph{Detection}
We capture the light {that returns} to the camera using an intensity array sensor (CMOS sensor) equipped with a QWP and a LP in front of the camera.
The QWP and the LP are mounted on rotation stages at angles $\theta_k^3$ and $\theta_k^4$, resulting in the light tensor arriving at the pixel $s$ on the CMOS sensor
\begin{equation}\label{eq:detection_mueller_transient}
\mathcal{I}^{s',\theta_k^4,\theta_k^3,\theta_k^2,\theta_k^1}\left(s,:\right) = \mathbf{L}^{\theta_k^4}\mathbf{Q}^{\theta_k^3}\mathcal{R}^{\theta_k^2,\theta_k^1}(s,:).
\end{equation}
For a 2D CMOS sensor pixel $s$, our spatio-polarimetric measurement $f_{sp}$ is then formed as the total recorded radiance of incident light, that is
\begin{equation}\label{eq:intensity_mueller_transient}
f_{sp}(\theta_k^4,\theta_k^3,\theta_k^2,\theta_k^1, s, s')=\mathcal{I}^{s',\theta_k^4,\theta_k^3,\theta_k^2,\theta_k^1}\left(s,0\right).
\end{equation}

\paragraph{Rotating Polarimetric Modulation}
The rotation angles of the two LPs and the two QWPs are crucial for diverse polarimetric sampling.
We evaluate two configurations including the established DRR scheme implemented with the fixed LPs and rotating QWPs at manually-chosen angles~\cite{azzam1978photopolarimetric} and our learned QRPR method with rotating LPs and QWPs at learned angles, see Figure~\ref{fig:ellipsometry_opt} for the specific angles we used.

\paragraph{Polarizer-array Sensor}
By replacing the CMOS intensity camera with a polarizer-array camera, we can sample multiple linear polarization states simultaneously, further reducing the number of captures as described in Section~\ref{sec:learned_ellipsometry}.
We experimentally implemented this polarizer-array configuration shown in Figure~\ref{fig:imaging}.

\begin{figure*}[t]
  \centering
  \includegraphics[width=\linewidth]{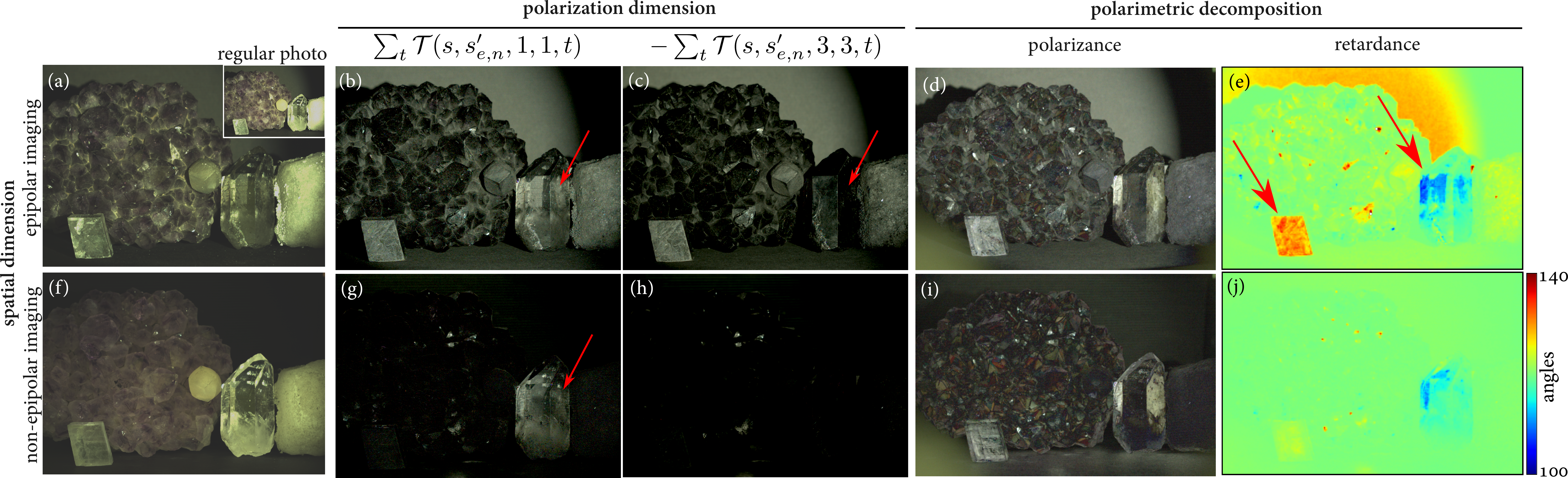}
  \caption{\label{fig:sp_light_transport}
  A conventional photo integrates light intensity reflected from a scene over temporal and spatial dimensions.
  {Specifically, conventional photography captures $\sum_{t,s'}{\mathcal{T}(s,s',0,0,t)}$, while losing} geometric and material information encoded in the scene transport.
  Spatially resolving the light transport allows for (a) epipolar and (f) {non-epipolar} imaging that approximates direct and indirect light transport components respectively as $\sum_{t}{\mathcal{T}(s,s'_e,0,0,t)}$ and $\sum_{t}{\mathcal{T}(s,s'_n,0,0,t)}$.
  Our spatio-polarimetric imaging approach acquires the complete polarimetric scene response which reveals strongly material-dependent structure in the transport.
  We observe the specular components {that maintain} horizontal linear polarization in (b) the approximated direct and (g) indirect reflection by epipolar and {non-epipolar} imaging.
  Note that the scattering inside of the crystal can be observed in the {non-epipolar} polarimetric component (g).
  (c) Analyzing the epipolar circular-polarization component, we can examine the surface of the transparent crystal.
  We further decompose the spatio-polarimetric light transport to analyze (d) \& (i) polarizance and (e) \& (j) retardance of the crystal.
  This reveals the birefringence properties of crystals indicated by red arrows. The two indicated crystal objects are made of calcite (left) and quartz (right) respectively.
  }
\end{figure*}
\subsection{Coaxial Temporal-Polarimetric Imaging}
Next, we describe the second instance of the proposed light transport theory.
Specifically, to acquire temporal-polarimetric light transport $\mathcal{T}_{pt}$, we present a novel coaxial ellipsometric SPAD-laser imaging method.
The proposed system consists of an illumination module and a detection module; a pulse of light is emitted to a scene and detected by a high-speed detector SPAD where both emission and detection paths involve polarization-modulating optics.
See Figure~\ref{fig:imaging}(b) for the schematic diagram and prototype photograph.

\paragraph{Illumination}
We use a picosecond pulsed laser to emit a linearly-polarized pulse of light at time zero at the linear polarization angle of zero $\theta_k^1=0^\circ$.
The polarization state of the emitted pulsed light is then modulated by a QWP mounted on a rotation stage at an angle $\theta_k^2$.
Light passes through a non-polarizing beam splitter in transmission mode and is reflected by a 2D galvo mirror specified by a spatial orientation $s$, exiting the imaging system towards a scene.
Then, for each QWP angle $\theta_k^2$ and spatial orientation $s$, the light emitted to a scene is modeled as a tensor
\begin{equation}\label{eq:illumination_stokes_transient}
\mathcal{P}^{\theta_k^2,\theta_k^1}(s,:) = \mathbf{G}^s\mathbf{B}_t\mathbf{Q}^{\theta_k}\mathbf{p},
\end{equation}
where $\mathbf{p}$ is the Stokes vector of the linearly polarized laser pulse at the polarization angle of $\theta_k^1=0^\circ$,
$\mathbf{B}_t$ is the beamsplitter Mueller matrix in transmission mode,
$\mathbf{G}^s$ is the galvo mirror Mueller matrix oriented to the spatial direction $s$.

\paragraph{Temporal-Polarimetric Light Transport}
The emitted light travels through a scene and arrives at a detection module after potentially complex light-matter interactions.
This temporal-polarimetric light transport is modeled as a tensor $\mathcal{T}_{pt}$ and is applied to the illumination tensor $\mathcal{P}^{\theta_k}$, resulting in the reflected tensor $\mathcal{R}^{\theta_k}$ as
\begin{equation}\label{eq:temporal_polarimetric_light_transport}
\mathcal{R}^{\theta_k^2,\theta_k^1}(s,:,t) = \mathcal{T}_{pt}(s,s,:,:,t)\mathcal{P}^{\theta_k^2,\theta_k^1}(s,:).
\end{equation}
To improve signal-to-noise ratio (SNR), we design our imaging system {to be coaxial with} the non-polarizing beamsplitter, sharing the {same} optical path for detection and emission.
Therefore, we use the same spatial sample $s$ for the illumination and the detection.

\paragraph{Detection}
The light returning from the scene first passes through the 2D galvo mirror.
It then enters the non-polarizing beamsplitter with reflection mode followed by two polarizing optics, a QWP and a LP.
The QWP and the LP are mounted on rotation stages at angles $\theta_k^3$ and $\theta_k^4$.
The SPAD detector captures the temporal response of the scene at a specific spatial orientation $s$ as
\begin{equation}\label{eq:detection_mueller_transient}
\mathcal{I}^{\theta_k^4, \theta_k^3,\theta_k^2, \theta_k^1}\left(s,:,t\right) = \mathbf{L}^{\theta_k^4}\mathbf{Q}^{\theta_k^3}\mathbf{B}_r\mathbf{G}^s\mathcal{R}^{\theta_k^2,\theta_k^1}(s,:,t),
\end{equation}
where $\mathbf{B}_r$ is the Mueller matrix of the beamsplitter in reflection mode.
Then, the SPAD sensor measures the total radiance and we
write this temporal-polarimetric image formation $f_{pt}$ as
\begin{equation}\label{eq:intensity_mueller_transient}
f_{pt}(\theta_k^4, \theta_k^3,\theta_k^2, \theta_k^1, s, t) = \mathcal{I}^{\theta_k^4, \theta_k^3,\theta_k^2, \theta_k^1}\left(s,0,t\right).
\end{equation}
For our temporal-polarimetric experiments, we fix the LP angles as zero.
We refer to the Supplemental Document for discussions on extending our quad-rotating ellipsometry to this temporal setup.

\section{Polarimetric Reconstruction}
\label{sec:recon}
We record spatial and temporal information as \emph{intensity} observations using a 2D CMOS camera and a SPAD sensor combined with a time-correlated single photon counting (TCSPC) device.
As these conventional sensors cannot directly measure the polarization state of light, we computationally recover polarimetric light transport in the form of Mueller matrix reconstruction from intensities observed with varying angles of polarizing optics.
Note that the Mueller matrix corresponds to the polarimetric slicing of the full light transport tensor.

\begin{figure*}[t]
  \centering
  \includegraphics[width=0.95\linewidth]{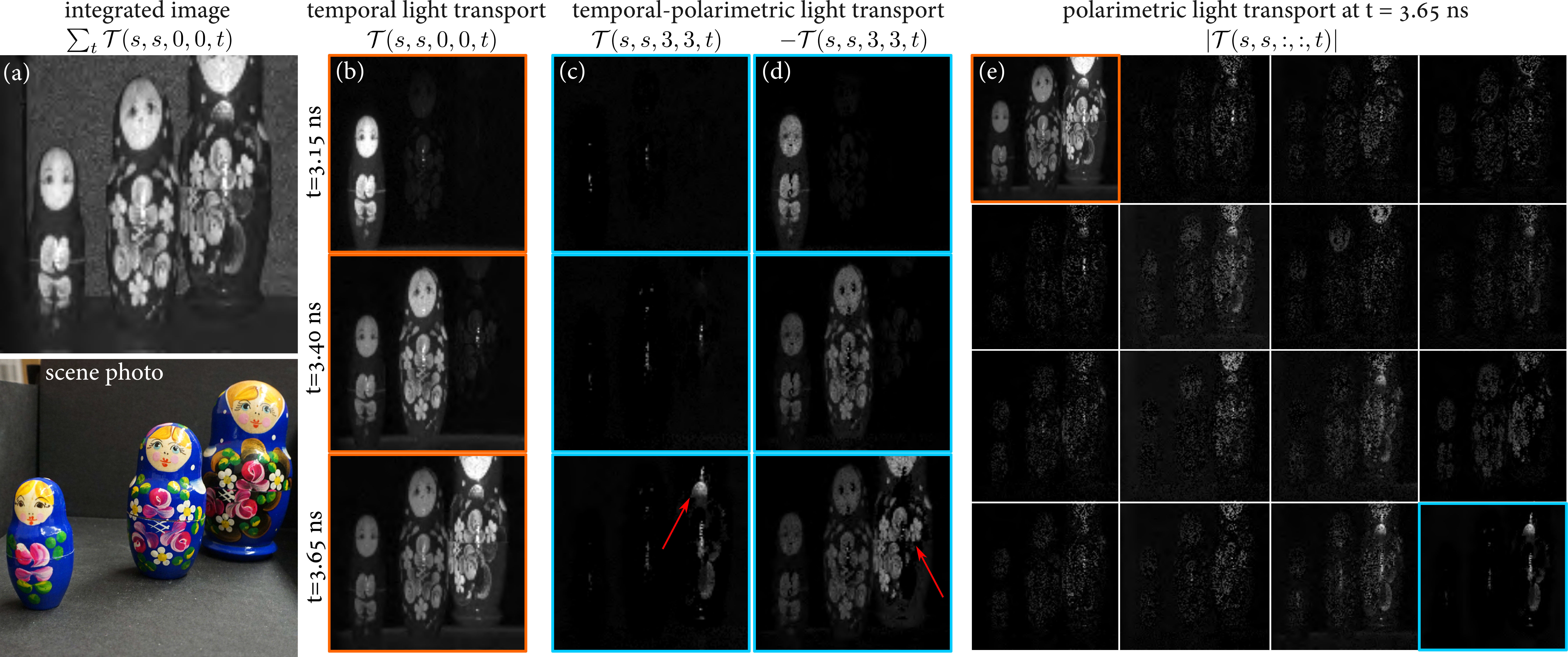}
  \caption{\label{fig:tp_light_transport}
  (a) Coaxial monochromatic imagers integrates temporal dimension without polarimetric resolution, capturing only intensity of light $\sum_{t}{\mathcal{T}(s,s,0,0,t)}$.
  (b) Transient imaging disentangles the time integration enabling light-in-flight imaging ${\mathcal{T}(s,s,0,0,t)}$.
  (e) Our coaxial temporal-polarimetric imaging method captures not only incoming intensity, but the complete polarization change at the picosecond temporal resolution as ${\mathcal{T}(s,s,:,:,t)}$.
  We visualize the $4\times4$ temporal-polarimetric light transport of the travel time 3.65\,ns. This temporal-polarimetric decomposition enables us to see (c) strong specular reflection that preserves the circular polarization state of incident light as ${\mathcal{T}(s,s,:,:,t)}$ and (d) specular components that invert the circular polarization of light similar to mirror as  $-{\mathcal{T}(s,s,:,:,t)}$. This reveals the material difference of the hand-painted patterns on the doll as indicated by the red arrows.
  }
\end{figure*}

To recover polarimetric information, we record a set of intensities $I^{\Theta_k}(s,t)$ and $I^{s',\Theta_k}(s)$ with varying angle of the polarizing optics, where $\Theta_k$ describes the angle configuration as $\{\theta_k^4,\theta_k^3,\theta_k^2,\theta_k^1\}$.
From the polarization-modulated observations, we reconstruct the polarimetric light transport $\mathbf{T}_p$, a slice of the full light transport tensor $\mathcal{T}$, by solving a least-squares optimization problem.
For spatio-polarimetric imaging, this is
\begin{equation}\label{eq:recon_mueller_spatial}
\mathcal{T}_{sp}(s,s',:,:)=\mathop {\arg \min }\limits_{\mathbf{T}_p} \sum\limits_{k = 1}^{K} {\left( I^{\Theta_k}(s,s') - f_{sp}(\Theta_k,s,s')\right)}^2,
\end{equation}
and for temporal-polarimetric imaging we solve
\begin{equation}\label{eq:recon_mueller_transient}
\mathcal{T}_{pt}(s,s,:,:,t)=\mathop {\arg \min }\limits_{\mathbf{T}_p} \sum\limits_{k = 1}^{K} {\left( I^{\Theta_k}(s,t) - f_{pt}(\Theta_k,s,t)\right)}^2.
\end{equation}
Specifically, we solve these problems {per pixel} with the least-squares solver inspired by {Baek et al.~\shortcite{baek2020image}} but with a different image formation model, see Supplemental Document.
The polarimetric recovery completes the proposed acquisition of the spatio-polarimetric transport tensor $\mathcal{T}_{sp}$ and the temporal-polarimetric transport tensor $\mathcal{T}_{pt}$.
See Figure~\ref{fig:sp_light_transport} and Figure~\ref{fig:tp_light_transport} for the reconstruction results.

\subsection{Polarimetric Decomposition}
We further analyze the polarimetric light transport for the basic polarization properties of a material: depolarization, retardance, and diattenuation.
These three polarimetric properties have been traditionally studied in optics~\cite{lu1996interpretation} and other application domains, notably biology~\cite{ghosh2011tissue}.
We bring these concepts to graphics and vision with an application to light transport analysis.
Depolarization describes how much light loses its polarization state, turning into unpolarized light for both linear polarization and circular polarization. As such, this quantity can be used to characterize {properties of scattering media} such as density and particle size.
Retardance describes the phase delay between two orthogonal polarization states including vertical/horizontal linear polarization and right/left circular polarization, therefore revealing the difference of the polarization-dependent refractive index of a material.
Retardance has been used to examine plastic stretch, defects on thin films, and birefringent materials.
As a generalized concept of dichroism, diattenuation describes the differential attenuation of light's orthogonal polarization states including linear and circular polarization.
Representing the three basic properties of depolarization, retardance, and diattenuation as individual Mueller matrices $\mathbf{M}_{\Delta}$, $\mathbf{M}_{R}$, and $\mathbf{M}_{D}$, we decompose the polarimetric light transport $\mathbf{T}_p$ into a product of three Mueller matrices following the optical Lu-Chipman decomposition~\shortcite{lu1996interpretation}
\begin{equation}\label{eq:polar_decomp}
  \mathbf{T}_p = \mathbf{M}_{\Delta} \mathbf{M}_{R} \mathbf{M}_{D}.
\end{equation}
Once decomposed, we analyze the low-dimensional projections of the Mueller matrices as three scalar quantities of polarizance, retardance, and diattenuation for intuitive analysis~\cite{ghosh2011tissue}.
Polarizance $P$ is the degree of polarization of exiting light after light-matter interactions of an unpolarized incident light in a scene as
\begin{equation}\label{eq:polarizance}
  P = \frac{1}{\mathbf{T}_p(0,0)}\sqrt{\mathbf{T}_p^2(1,0)+\mathbf{T}_p^2(2,0)+\mathbf{T}_p^2(3,0)}.
\end{equation}
Retardance $R$ is the phase shift between {the} orthogonal polarization states of light
\begin{equation}\label{eq:retardance}
  R = \cos^{-1}(\frac{\mathrm{tr}(\mathbf{M}_R)}{2} - 1),
\end{equation}
where $\mathrm{tr}(\cdot)$ is the trace operator.
Diattenuation $D$ characterizes how incident light polarization affects the transmission of light
\begin{equation}\label{eq:diattenuation}
  D = \frac{1}{\mathbf{T}_p(0,0)}\sqrt{\mathbf{T}_p^2(0,1)+\mathbf{T}_p^2(0,2)+\mathbf{T}_p^2(0,3)}.
\end{equation}
See Figure~\ref{fig:sp_light_transport} for decomposition.

\section{Experimental Prototype}
We describe our two experimental prototypes for spatio-polarimetric and coaxial temporal-polarimetric acquisition.
See Figure~\ref{fig:imaging} for schematic diagrams and system photographs.
For details, including calibration procedures, we refer to {the} Supplemental Document.

\subsection{Spatio-polarimetric Setup}
\paragraph{Spatial Imaging}
To acquire spatially-resolved light transport, we use a DMD projector (Lightcraft LC3000-G2-Pro) and a machine-vision camera (FLIR GS3-U3-51S5C) with an objective lens of focal length 25\,mm (Fujinon HF25HA-1S 25mm f/1.4). The DMD projector emits unpolarized light with a programmable projection pattern $M_p(s')$ and the camera has a programmable region-of-interest feature to implement the imaging pattern  $M_c(s)$. While arbitrary projection and imaging patterns can be implemented, we demonstrate two effective mask configurations for epipolar and {non-epipolar} imaging proposed in O'Toole et al.~\shortcite{o2012primal}. To this end, we geometrically calibrate the projector and the camera by capturing checkerboard images at different poses, obtaining the intrinsic and distortion parameters of the devices, and the fundamental matrix between them in the undistorted coordinates. This allows us to compute epipolar/{non-epipolar} patterns in the distorted coordinates that can be used as $M_p(s')$ and $M_c(s)$.

\begin{figure}[t]
  \centering
  \includegraphics[width=0.95\linewidth]{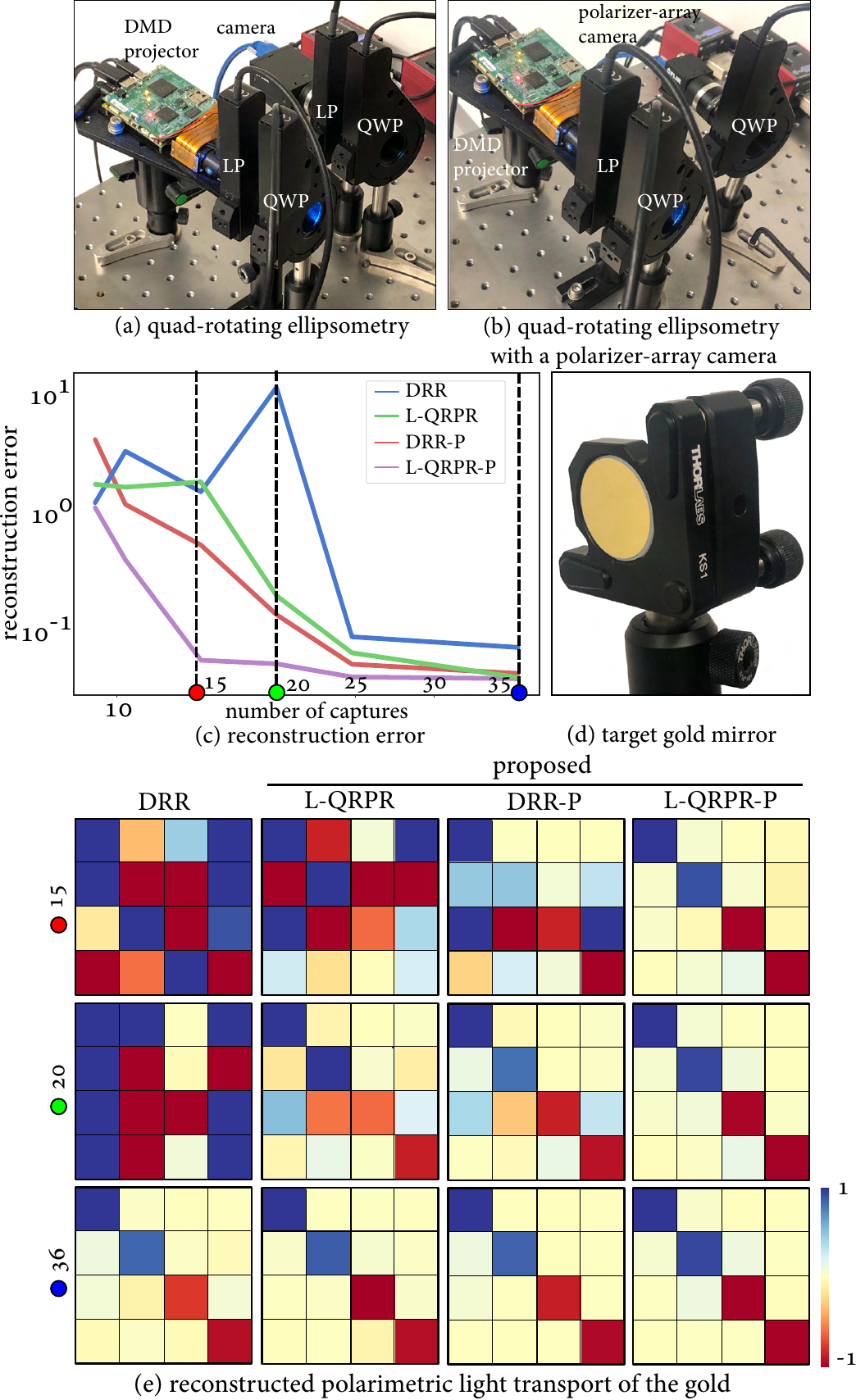}
  \vspace{-4mm}
  \caption{\label{fig:learned_ellipsometry_eval}
  We modify our spatio-polarimetric imager to demonstrate the proposed learned ellipsometry method of (a) the quad-rotating ellipsometry and (b) its extension with the polarizer-array sensor. (d) We capture the polarimetric light transport of {an} unprotected gold mirror. (c) \& (e) Our method outperforms the conventional DRR method, achieving {same reconstruction accuracy with fewer captures}. We use acronyms L-QRPR for the learned quad-rotating polarizer retarder, DRR-P for the dual-rotating retarder with a polarizer sensor, and L-QRPR-P for the learned quad-rotating polarizer retarder with a polarizer sensor. Note that compared to the conventional DRR method, we reduce the number of captures from 36 to 15 for the L-QRPR-P method.
  }
  \vspace{-5mm}
\end{figure}

\paragraph{Polarimetric Imaging}
We use a pair of achromatic QWPs (Thorlabs AQWP10M-580) and LPs (Newport 10LP-VIS-B) in front of the projector and the camera.
Unpolarized light emitted from the DMD projector passes through a LP mounted on a manual rotary stage (Newport RM25A) {along a} vertical axis for the DRR ellipsometry.
For our learned rotating ellipsometry, the LP is mounted on a motorized rotary stage (Thorlabs KPRM1E).
The linearly polarized light is then modulated by a QWP mounted on another motorized rotary stage so that the fast axis is aligned with the polarization axis of the LP at the zero base.
The second module of a QWP and a LP is installed in front of the camera, modulating the polarization state of the returning light from a scene.
For DRR ellipsometry, we rotate the QWPs in front of the projector and the camera 36 times at angles $\theta_k=5k$ and $\theta'_k=25k,$ where $k\in \{0,...,35\}$.
For our learned ellipsometry, we use the optimized angles of the polarizing optics as shown in Figure~\ref{fig:ellipsometry_opt}.
{We obtain all the results with the DRR setting except for the evaluation experiments for the learned ellipsometry (Figures~\ref{fig:learned_ellipsometry_eval} and \ref{fig:learned_ellipsometry_scene}).
}
For each angle index $k$, we obtain epipolar and {non-epipolar} images with the precomputed spatial masks $M_p$ and $M_c$.

\paragraph{Acquisition Parameters}
We set the acquisition parameters to obtain accurate reconstructions of light transport tensors at acceptable acquisition time.
Our camera is set to have 600\,ms exposure and captures 16bit raw images.
The projector is configured to emit a 8bit intensity image and we use zero or 255 depending on whether the projector mask is turned off or on.
For each polarizing-optics angle set $\Theta_k=\{\theta_k^1,\theta_k^2,\theta_k^3,\theta_k^4\}$, we project a sequence of the precomputed masks for epipolar and {non-epipolar} imaging.
To accelerate the capture speed, we reconfigure the region of interest for the camera depending on the camera mask. In addition, the capture script is written in Python with multi-threading programming to reduce any capture-to-save delay.
This results in {an} acquisition time of one hour for the epipolar and {non-epipolar} polarimetric captures with DRR ellipsometry.

\subsection{Coaxial Temporal-polarimetric Setup}
\paragraph{Temporal Dimension}
To achieve high temporal resolution, we use a picosecond pulsed laser unit (Edinburgh instruments EPL-635) with {an} optical wavelength of 635\,nm.
The laser pulse is linearly polarized and passes through a non-polarizing beamsplitter (Thorlabs BS013) and is directed to a scene by {using} a mirror galvanometer (Thorlabs GVS012).
As our system is coaxial, we detect light coming from a scene using the same path by the non-polarizing beamsplitter.
A single-photon avalanche diode (MPD series SPAD) is used with a time-correlated single photon counter (PicoQuant TimeHarp 260 PICO) {for} controlling the timings of the SPAD and the laser.
Overall, our system has {a} temporal resolution of 25 picoseconds.

\paragraph{Polarimetric Dimension}
Similar to the spatio-polarimetric setup, we employ two QWPs and a LP.
Note that the laser light is linearly polarized to the vertical orientation and passes through the QWP mounted on a motorized rotation stage before entering the beamsplitter.
The same configuration of the QWP and the LP is installed for the detection path after the beamsplitter reflection.
{
We use the DRR configuration for all temporal-polarimetric experiments.
}

\paragraph{Acquisition Parameters}
We use {an} exposure time of 10\,ms for each galvo mirror position, which we sample at 128$\times$128 points.
We repeat this for every angle configuration of the polarizing optics {$\Theta_{k\in\{1,\cdots,36\}}$}.
{For a single scene, the total acquisition time is 7 hours.}

\section{Assessment}
\label{sec:results}
We experimentally validate the proposed method by acquiring and analyzing spatio-temporal polarimetric transport tensors.
For the first time, we demonstrate complete polarimetric light transport capture combined with temporal and spatial transport imaging. The resulting transport tensors provide detailed information about a scene both in terms of geometry and material composition, which are challenging to acquire for existing temporal and spatial transport methods.
We also experimentally validate our learned rotating ellipsometry and its polarizer-array variant.

\subsection{Spatio-Polarimetric Light Transport Analysis}
Spatial light transport methods are capable of isolating light paths satisfying a certain transport geometry connecting a projector pixel and a detector pixel.
Existing methods untangle light transport into different path geometries by masking transport components in the illumination path, using a projector, and on the sensor path, with per-pixel masks on a camera. Commonly used mask patterns include high-frequency checkerboards for direct-global separation~\cite{nayar2006fast} and epipolar/{non-epipolar} imaging configurations~\cite{o2012primal}. Although the proposed projector-camera ellipsometry method allows for arbitrary mask patterns, we demonstrate epipolar and non-epipolar imaging with complete polarimetric sampling.
As such, for the first time, we demonstrate the complete polarimetric decomposition of epipolar and non-epipolar light transport.
Figure~\ref{fig:sp_light_transport} shows the decomposed spatio-polarimetric light transport of crystals that reveal direct and indirect transport further resolved in polarization state.
This allows us to acquire specular reflections that maintain the horizontal linear polarization of incident light for direct and indirect reflections in $\sum_{t}{\mathcal{T}(s,s'_e,1,1,t)}$ and $\sum_{t}{\mathcal{T}(s,s'_n,1,1,t)}$.
Direct surface reflection on the crystals are revealed in the circular-polarization inverting component $-\sum_{t}{\mathcal{T}(s,s'_e,3,3,t)}$ with epipolar scanning that could allow for examining the surface profiles of the transparent crystal.
Further, we decomposed the epipolar/{non-epipolar} polarimetric light transport to obtain diattenuation, polarizance, and retardance.
This reveals the intrinsic properties of various crystals, notably the birefringence of calcite and quartz in the retardance estimates.
Note that complete Mueller matrix acquisition with spatial probing is essential for this spatio{-}polarimetric decomposition and its application to estimating crystal properties.

\subsection{Temporal-Polarimetric Light Transport Analysis}
Existing transient imaging methods capture light propagating through a scene at high temporal resolution on the order of picoseconds~\cite{velten2013femto,heide2013low,o2017reconstructing}.
This acquisition approach reveals the scene geometry both in the direct sight~\cite{heide2018sub} and outside the direct line of sight~\cite{velten2012recovering}.
However, material-dependent scene understanding has been challenging for transient imaging due to temporal blur and temporal metamerism that light rays undergo since different material-dependent scattering events have similar travel distances~\cite{wu2014decomposing}.
For the first time, we demonstrate that polarimetric transient imaging improves the decomposition capability of material-dependent light transport.
Figure~\ref{fig:tp_light_transport} shows the decomposed coaxial temporal-polarimetric light transport.
For each time bin $t$, we obtain {a} complete polarimetric decomposition.
{This allows us to separate specular reflections that maintaining circular polarization of incident light.}
Furthermore, we distinguish the specular components that invert or maintain the incident circular polarization with ${\mathcal{T}(s,s,:,:,t)}$ and $-{\mathcal{T}(s,s,:,:,t)}$.

\begin{figure}[t]
  \centering
  \includegraphics[width=\linewidth]{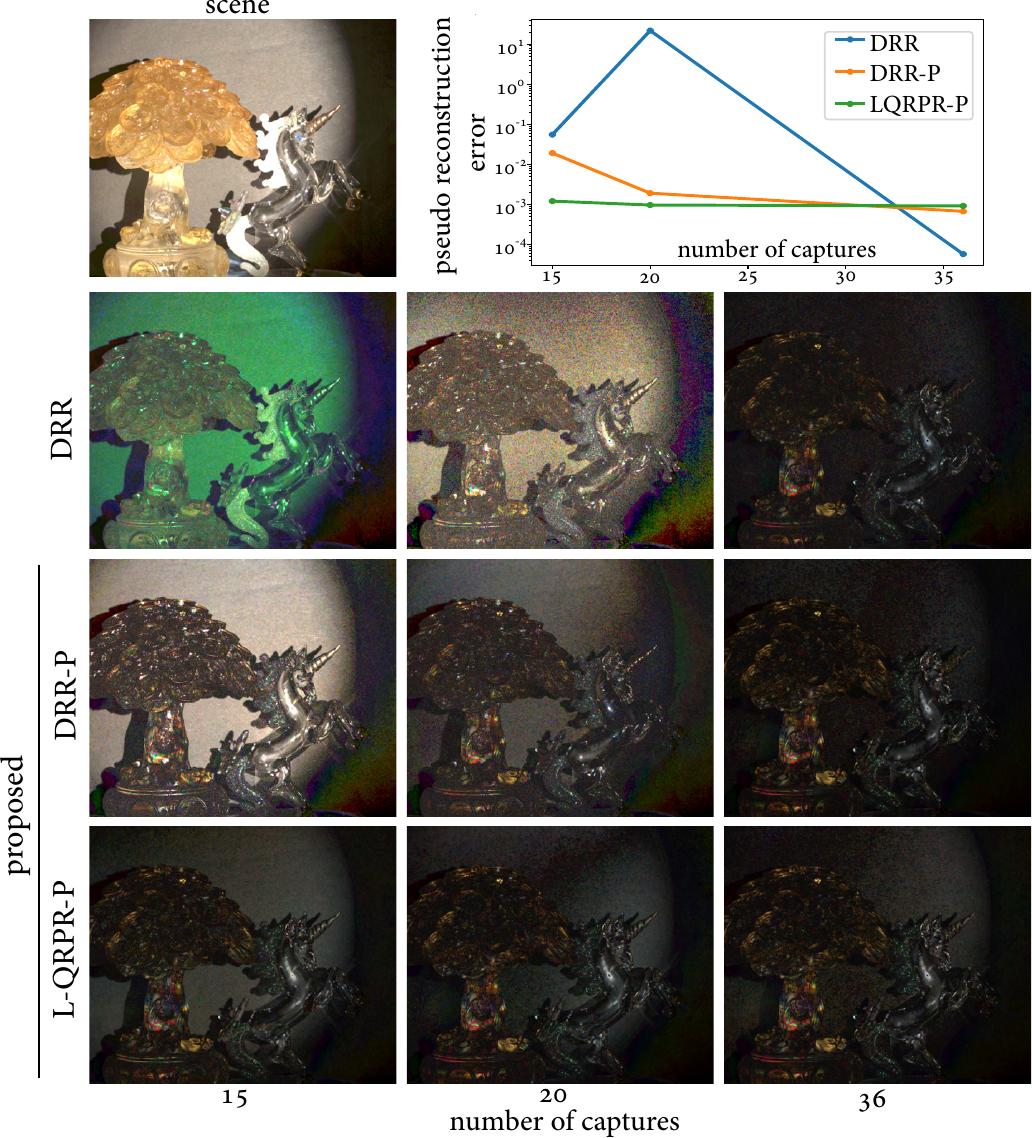}
  \vspace{-5mm}
  \caption{\label{fig:learned_ellipsometry_scene}
  {The proposed learned ellipsometry method efficiently acquires complex polarimetric light transport with fewer captures than existing methods.
  For quantitative evaluation, we use pseudo ground-truth data from 50 DRR captures~\cite{azzam1978photopolarimetric}. The proposed L-QRPR-P method provides accurate reconstructions despite only using 15 captures.
  We visualize the (2,3)-th component of the polarimetric transport $-\sum_{s',t}{\mathcal{T}(s,s',2,3,t)}$.}
  }
  \vspace{-6mm}
\end{figure}

\begin{figure*}[t]
  \centering
  \includegraphics[width=\linewidth]{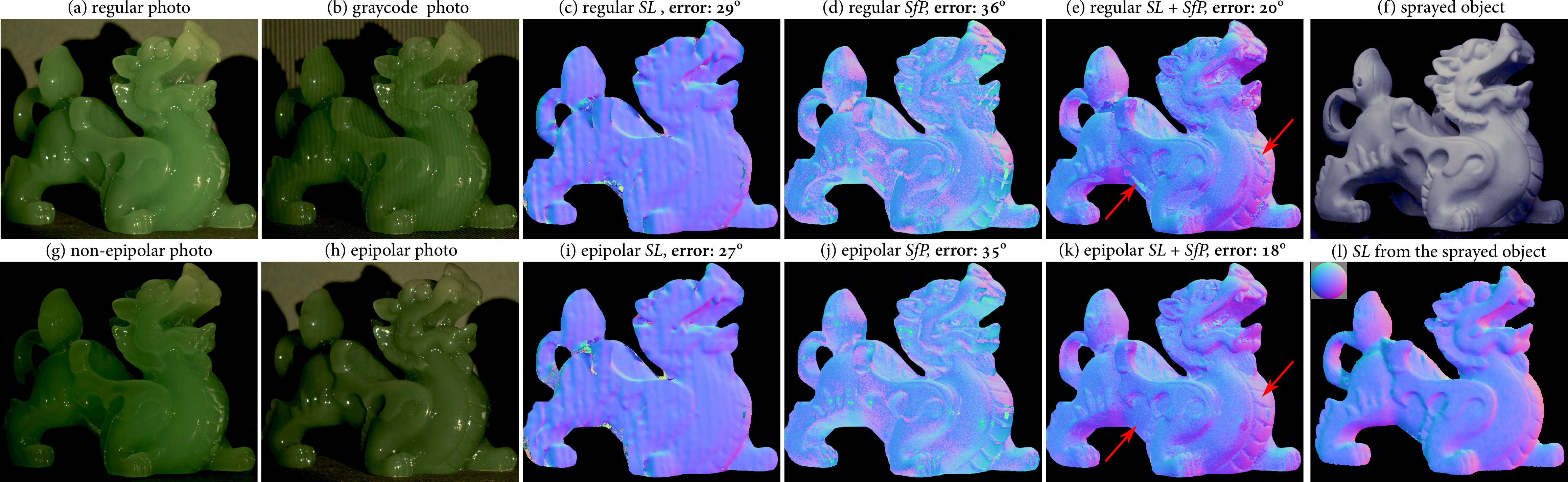}
  \caption{\label{fig:shape_reconstruction}
  (a) We reconstruct the shape of a dragon statue with strong subsurface scattering and interreflections.
  (g) \& (h) Spatial light transport of non-epipolar and epipolar imaging reveals the approximate indirect and direct components.
  (c) Conventional structured light (\textit{SL}) methods, such as (b) graycode probing, assume diffuse appearance and struggle with strong subsurface scattering.
  (d) Shape-from-polarization (\textit{SfP}) methods estimate surface normals with ambiguity in azimuthal angle and refractive index of the material~\cite{atkinson2006recovery}.
  (e) Combining structured light with the shape-from-polarization method resolves the polarimetric-normal ambiguity but the interreflections and subsurface scattering corrupt estimated normals. See red arrows.
  (k) Exploiting both geometric cues with epipolar capture eliminates the ambiguity in existing methods and allows us to reconstruct fine geometry detail.
  (f) We sprayed anti-reflective diffuse particles to the dragon statue, obtaining (l) the pseudo ground-truth geometry by using structured light scanning. We compute the reconstruction errors of surface normals with this registered ground truth, reported in the subfigure caption headings, revealing that our method also quantitatively outperforms existing methods. We refer to the Supplemental Document for further results.
  }
  \vspace{-5mm}
\end{figure*}
\subsection{Learned Ellipsometry and Polarizer-array Sensor}
In addition to the synthetic assessment in Figure~\ref{fig:ellipsometry_opt}, we experimentally evaluate our proposed ellipsometry methods compared to the conventional DRR method~\cite{azzam1978photopolarimetric}.
To this end, we modify our spatio-polarimetric setup by mounting the linear polarizers on the motorized rotation stages to implement the quad-rotating ellipsometry. For the polarizer-array extension, we replace the CMOS camera with the polarizer-array camera (FLIR Blackfly S BFS-U3-51S5PC).
For details on geometric and polarimetric calibrations, we refer to the Supplemental Document.
We capture an unprotected gold mirror (Thorlabs PF10-03-M03) which has known polarimetric light transport as shown in Figure~\ref{fig:learned_ellipsometry_eval}.
We confirm that our proposed ellipsometry reduces the number of captures required for acquiring polarimetric light transport compared to the conventional DRR technique~\cite{azzam1978photopolarimetric}.
{
We also evaluate the ellipsometric methods (DRR, DRR-P, L-QRPR-P) on a scene with unseen materials, see Figure~\ref{fig:learned_ellipsometry_scene}.
To this end, we use the polarizer-array setup and obtain pseudo ground-truth data using 50 DRR captures~\cite{azzam1978photopolarimetric}.
Note that DRR inputs can be acquired by sampling the on-sensor polarizers oriented at zero degree, resulting in pixel-aligned inputs to the DRR-P and the L-QRPR-P methods.
We again observe that L-QRPR-P provides reliable reconstructions of polarimetric transport with only 15 captures while DRR fails.
}
We refer to the Supplemental {Document} for {qualitative evaluation of all ellipsometry methods and} details on the spatio-polarimetric acquisition with the proposed ellipsometry method.

\begin{figure}[t]
  \centering
  \includegraphics[width=\linewidth]{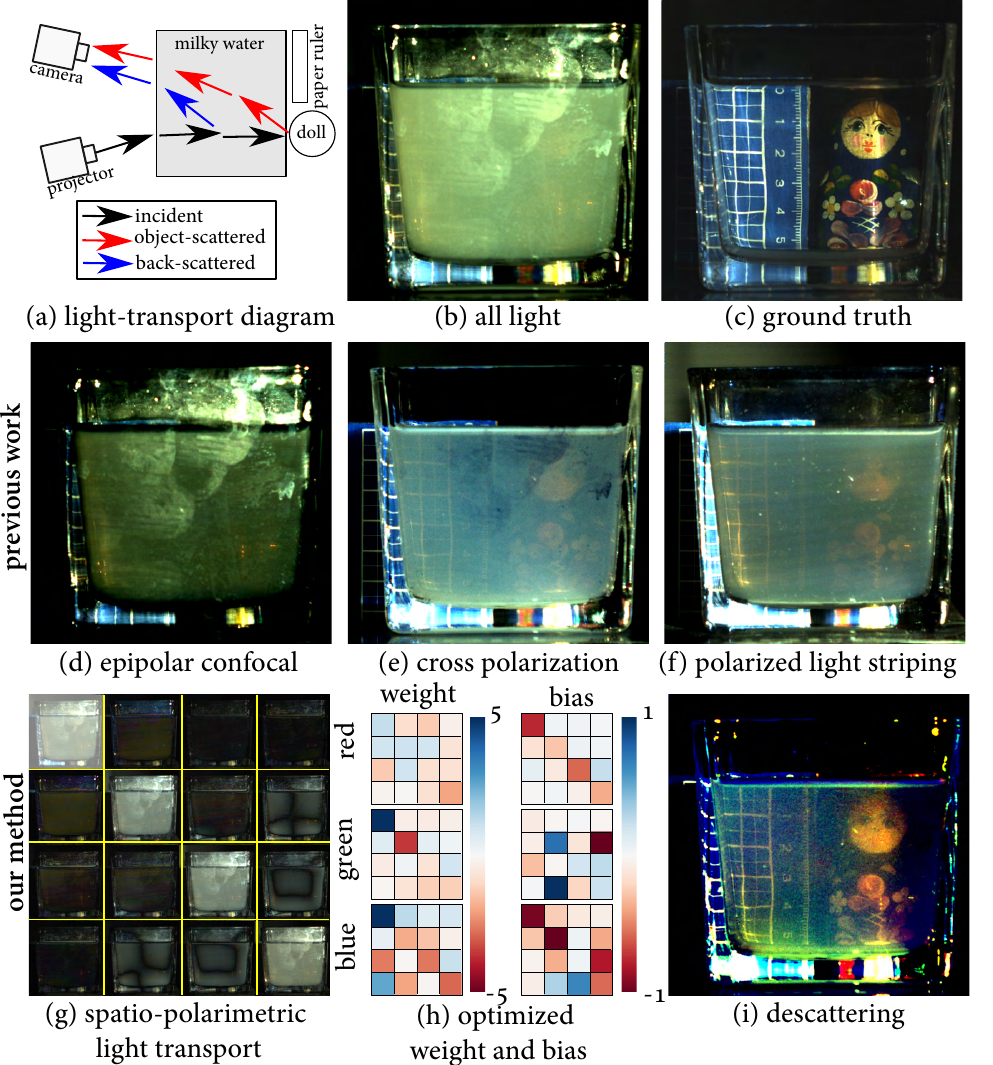}
  \caption{\label{fig:descattering_res}%
  Experimental seeing through scattering {media}.
  (a) Experimental configuration where milky water in a glass pot is in between the target objects and the imaging setup.
  (b) With severe backscatter, conventional imaging methods that integrate backscattered light and ballistic photons have limited visibility of (c) light interacting with the object.
  (d) Epipolar confocal imaging~\cite{o2012primal} removes scattering outside of the epipolar plane while epipolar scattering is ignored.
  (e) Cross-polarization imaging~\cite{treibitz2008active} relies on {the orthogonal polarization states} of linear/circular components for back-scattered and object-scattered light.
  (f) Polarized light striping~\cite{gupta2008controlling} combines line scanning and degree-of-polarization analysis on the linear polarization component, further improving descattering results.
  (g) We take a step forward by capturing the complete epipolar polarimetric light transport $\sum_{t} \mathcal{T}(s,s_e,:,:,t)$.
  (h) Our optimized weights and biases for polarimetric components reveal the contributions of all spatio-polarimetric information for descattering, leading to (i) the improved visibility of the hidden scene.
  }
  \vspace{-6.5mm}
\end{figure}

\section{Applications}
We demonstrate applications of the proposed method for shape reconstruction with subsurface scattering, seeing through scattering {media}, multi-bounce reflection analysis, and resolving metamerism.

\subsection{Shape Reconstruction with Subsurface Scattering}
Reconstructing the geometry of an object with strong subsurface scattering poses a challenge for existing 3D imaging systems.
The proposed spatio-polarimetric imaging method solves this problem by jointly exploiting polarimetric and spatial information that {encodes} information about a scene geometry. Figure~\ref{fig:shape_reconstruction} shows the reconstructed geometry for a dragon object with severe subsurface scattering and object interreflections.
We note that conventional structured-light methods, such as graycode patterns, assume diffuse surface properties, and, as such, suffer in presence of strong subsurface scattering and {interreflections} on the object, failing to resolve fine surface structure.
{Epipolar structured light methods approximately capture direct reflections} in order to mitigate such problems. However, epipolar structured light approaches \emph{fail with strong subsurface scattering} that does not fulfill the epipolar constraint.
Shape from polarization~\cite{atkinson2006recovery} provides geometric cues of the surface with $\pi$ azimuthal ambiguity.
{Although these cues allow for the recovery of surface normals for convex objects, interreflections can still cause severe errors for normal estimation.}
Even combining structured light and shape from polarization, which resolves the polarization-normal ambiguity, suffers from such interreflection artifacts~\cite{baek2018simultaneous}.
In contrast, we obtain robust geometry from structured-light normals and polarization normals using epipolar-polarimetric captures. As such, the proposed method is capable of exploiting the intertangled relationship between the epipolar spatial domain and polarization domain.
We observe improved geometric reconstruction compared to \cite{baek2018simultaneous} at the neck and torso of the dragon indicated by the red arrows in Figure~\ref{fig:shape_reconstruction}.
To quantitatively assess reconstruction quality, we have acquired the pseudo ground-truth geometry using structured light after spraying diffuse particles onto the object. In Figure~\ref{fig:shape_reconstruction}, we report the reconstruction errors of surface normals of each method with this {ground-truth} measurement after registration.
The proposed epipolar structured polarimetric scanning qualitatively and quantitatively outperforms existing methods as validated by the reported normal estimation error.
For details on the reconstruction method and additional results, we refer to the Supplemental Document.

\subsection{Seeing through Scattering {Media}}
Removing the effect of volumetric scattering is an open challenge in imaging and computer vision.
We demonstrate imaging through scattering {media} to see objects of interest through scattering {media} such as fog and milky water in macroscopic scenes.
In the presence of scattering in Figure~\ref{fig:descattering_res}(a), light transport $\mathcal{T}$ from a source to a detector is modeled as a sum of the light $\mathcal{T}_{b}$ back-scattered by the scattering and the ballistic object-scattered light $\mathcal{T}_{o}$ reflected on the object surfaces: $\mathcal{T}=\mathcal{T}_{b}+\mathcal{T}_{o}$. Descattering then refers to the process of recovering the object-scattered component $\mathcal{T}_{o}$ from the integrated transport $\mathcal{T}$. Existing methods have shown that the structure in polarimetric light transport can aid the recovery in this ill-posed problem.
A key intuition of existing approaches is that the back-scattered component $\mathcal{T}_{b}$ tends to maintain the polarization of incident light, while the object-scattered component $\mathcal{T}_{o}$ is depolarized~\cite{treibitz2008active,gupta2008controlling}. As such, capturing two images with parallel and orthogonal states of the linear polarizers {allows us to} filter out linearly polarized light which is assumed to be the backscattered light. While this approach reduces some backscatter, it relies on incomplete polarization analysis; only one linearly-polarized angle is investigated.
{Other linear polarization angles, circular,} and elliptical polarization are ignored. Existing methods also have explored spatial probing~\cite{fuchs2008combining,o2012primal,gupta2008controlling} for descattering but independently of the polarimetric analysis.

We overcome the limitations of existing approaches by imaging through scattering {media} with a variant of the proposed spatio-polarimetric imaging system. Our system allows us to capture the complete polarimetric change in scattering {media}, in contrast to the existing incomplete polarization-based descattering methods, while simultaneously performing epipolar probing. To recover the undisturbed latent image (without the scattering {media} present), we solve an optimization problem to estimate the optimal weight and bias of the spatio-polarimetric light transport for descattering, that is
\begin{align}\label{eq:descattering}
& \mathop {{\mathrm{minimize}}}\limits_{W,b} \left\| S_\mathrm{syn} - S_\mathrm{gt} \right\|_2, \nonumber \\
S_\mathrm{syn}^{k\in \{R,G,B\}} &=\sum\limits_{i = 0}^3 \sum\limits_{j = 0}^3 \left\{ W_{ijk} \odot \left( \sum_t\mathcal{T}_{k}(s,s_e,i,j,t) + b_{ijk} \right) \right\},
\end{align}
where $W$ and $b$ are the 4$\times$4 matrices for trichromatic channels globally applied to {all} pixels, $\mathcal{T}_k$ is the transport tensor at $k$-th color channel,  and $S_\mathrm{gt}$ is the ground-truth image without scattering.
We solve this optimization problem with a quasi-Newton L-BFGS optimizer~\cite{liu1989limited}.
Figure~\ref{fig:descattering_res} shows that our method provides improved visibility of the doll and the ruler compared to existing approaches.
It is interesting to note that the optimized weight and bias perform linear operations on all polarimetric elements including linear, circular, and elliptical components across the 4$\times$4 polarimetric Mueller matrix.
This shows that complete polarization sensing combined with the spatial epipolar scanning enables effective descattering.

\begin{figure*}[t]
  \centering
  \includegraphics[width=0.9\linewidth]{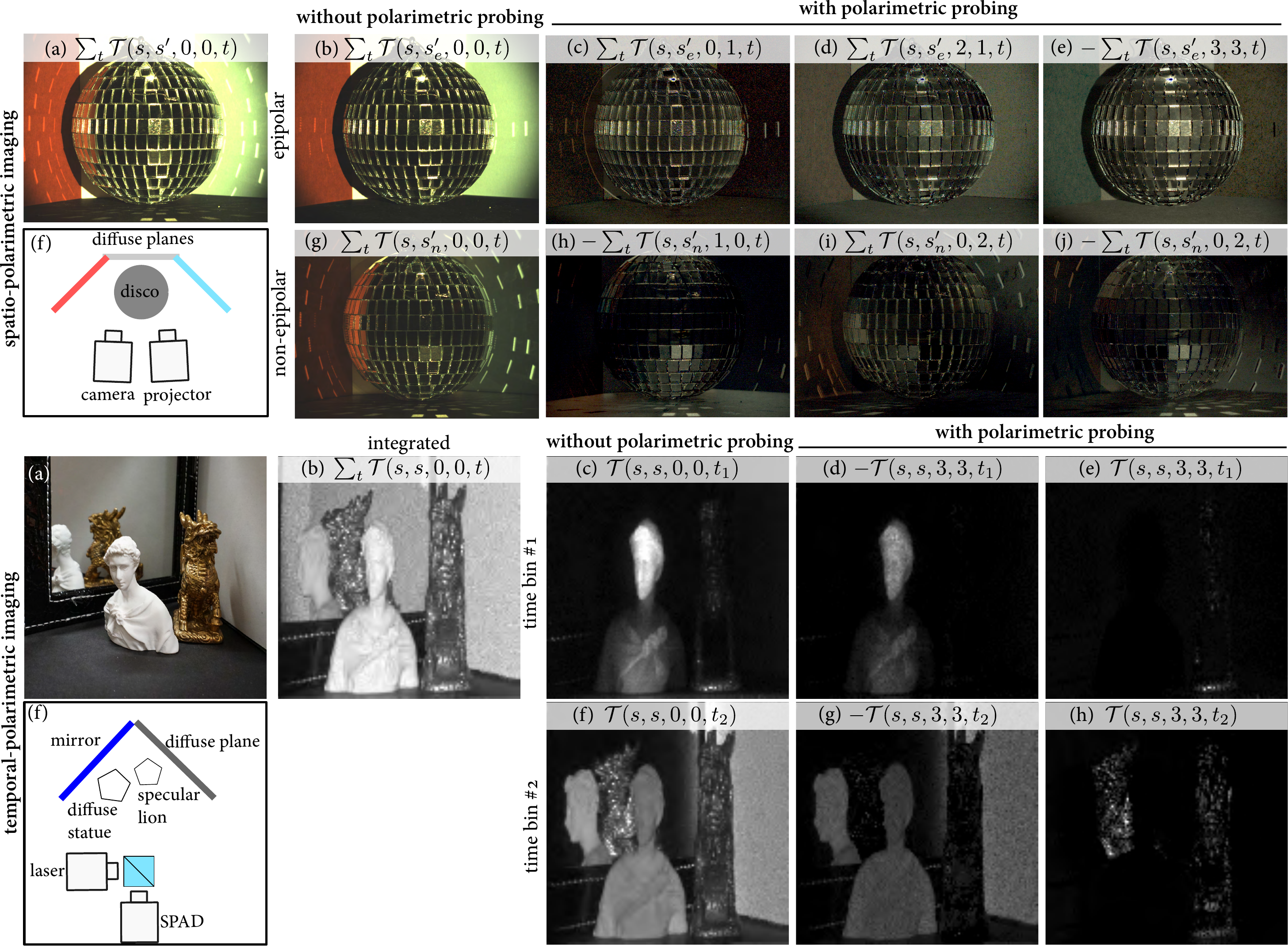}
  \caption{\label{fig:reflection_removal}
    {Spatio-polarimetric Imaging}: (a) \& (f) We untangle complex reflections between a specular disco ball and diffuse planes using our spatio-polarimetric imaging approach. (b) Epipolar and (g) {non-epipolar} spatial imaging partially separates direct and indirect reflections between the objects~\cite{o2012primal}. Our epipolar polarimetric probing (c) isolates the remaining specular reflection on the diffuse planes ($\sum_{t}{\mathcal{T}(s,s'_e,0,1,t)}$), (d) removes the specular reflection ($\sum_{t}{\mathcal{T}(s,s'_e,2,1,t)}$), (e) and captures the diffuse direct interreflections between the red and blue planes as revealed by the changed colors of the planes ($-\sum_{t}{\mathcal{T}(s,s'_e,3,3,t)}$).
    Our polarimetric non-epipolar imaging provides further decomposition of the indirect reflections, allowing us to untangle (h) the reflection from the planes and the disco ball only to the floor ($-\sum_{t}{\mathcal{T}(s,s'_n,1,0,t)}$), (i) \& (j) the reflections with different directions from the disco ball to the planes ($\sum_{t}{\mathcal{T}(s,s'_n,0,2,t)}$ and $-\sum_{t}{\mathcal{T}(s,s'_n,0,2,t)}$).
    {Temporal-polarimetric imaging}: (a) \& (b) Conventional photography integrates {along the} time dimension, capturing the steady-state equilibrium of light transport. (c) \& (f) Transient imaging captures the temporal dimension, revealing light in flight. Our temporal-polarimetric probing further decomposes this temporal transport depending on material properties, effectively distinguishing (d) \& (g) the diffuse transient reflections as ${\mathcal{T}(s,s,3,3,t_1)}$ and (e) \& (h) specular transient reflections from metallic surfaces as ${\mathcal{T}(s,s,3,3,t_2)}$.
    }
  \vspace{-4mm}
\end{figure*}

\begin{figure}[t]
  \centering
  \includegraphics[width=\linewidth]{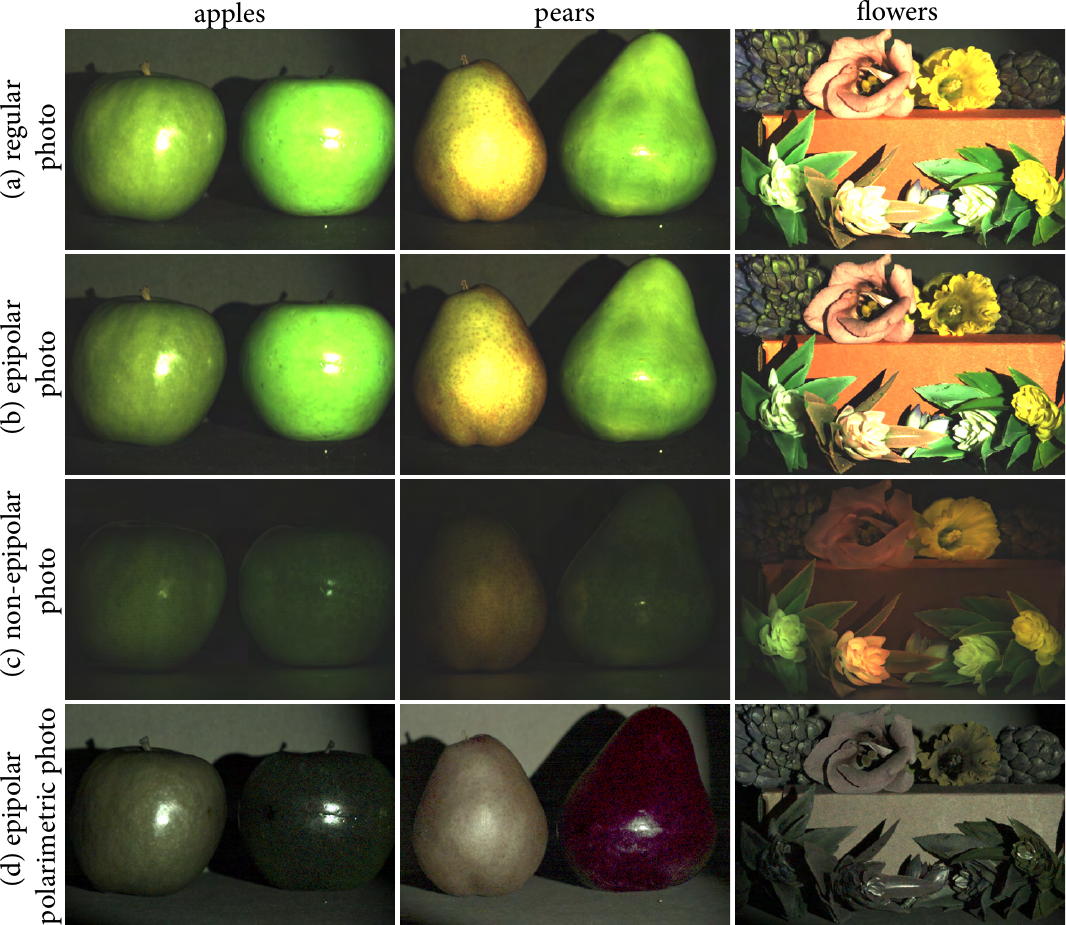}
  \caption{\label{fig:material_analysis}
   The proposed spatio-polarimetric imaging approach allows us to resolve metamerism observed in (a) real and fake fruits and flowers.
   (b) \& (c) Conventional spatial epipolar and {non-epipolar} imaging methods cannot identify the subtle difference between the real and the fake.
   (d) Our spatio-polarimetric imaging reveals that the fake plastic objects show minimal subsurface scattering. The right apple, right pear, and the bottom flowers are fake objects. These objects contain low-intensity polarimetric scattering effects as shown in the epipolar 45-degree linear-polarization component $-\sum_t{\mathcal{T}(s,s'_e,2,2,t)}$.
   }
   \vspace{-7mm}
\end{figure}

\subsection{Untangling Multi-Bounce Light Transport}
The light transport in a complex scene featuring diverse materials can include multiple interreflection between individual scene surface points. In conventional intensity images, the superposition of this multi-bounce transport results in complex {diffusion} and specular caustics that are challenging to analyze {with conventional intensity images}. We demonstrate that adding spatio-polarimetric and temporal-polarimetric {separation resolves} this ambiguity.
Figure~\ref{fig:reflection_removal} shows that we can {decouple interreflections between the specular coins, diffuse floor, specular disco ball, and colored diffuse planes.}
By exploiting the complicated spatio-polarimetric transport response, we separate the specular interreflections between the coins and diffuse interreflection between the colored diffuse planes with the disco ball.
This can be explained by the fact that specular interreflection often flips the circular polarization, while diffuse interreflection is encoded in the linear polarization.
Also, we separate reflections from the statue/lion and the mirror with specular and diffuse decomposition.
This fine-grained decomposition of light transport is enabled by our joint analysis of spatial and polarimetric light transport.

\subsection{Breaking Metamerism with Polarization}
The proposed polarized light transport analysis in the temporal and spatial dimensions {allows} us to resolve metamerism. Conventional cameras integrate over large regions of the spectral irradiance, which can make objects with different reflectance appear identical as a result of this integration. Acquiring polarimetric light transport allows us to resolve this ambiguity without having to acquire hyperspectral data with high spectral resolution. Figure~\ref{fig:material_analysis} validates that the proposed polarimetric analysis allows us to identify different materials for various objects including fake/real fruits and flowers by discovering scattering that {are} ``hidden'' in conventional and spatial captures.
{Here} we utilize the epipolar spatial dimension with 45 degree linear-polarization maintaining component, $-\sum_t{\mathcal{T}(s,s'_e,2,2,t)}$, where $s$ and $s'_e$ are the pairs of pixels in the camera and the projector within the same epipolar plane.
In this tensor slice, the plastic fake objects of fruits and flowers have low intensity because the captured light returns from the direct surface reflection from the surface paint. In constrast, the real organic materials present rich subsurface scattering revealing its unique organic material structure.

\section{Discussion}
In this work, we develop a method for capturing a light transport tensor decomposed into its spatial, temporal, and polarimetric dimensions. This comes at the cost of sequential acquisition and reduced light efficiency. We partly mitigated the light efficiency issue in our spatio-polarimetric imaging prototype by spatially binning 2$\times$2 pixels, {resulting in an effective pixel pitch of 12.9$\mu$m}. Departing from sequential acquisition and developing photon-efficient capture are exciting areas of future research.
{In particular, one potential direction that we envision is to combine the proposed method with liquid crystal retarders which could substantially reduce capture time, possible even below a minute.} Our current implementation of the polarimetric reconstruction requires 12 sec. and 0.3 sec. to process the spatio-polarimetric and the temporal-polarimetric data, respectively. Accelerating the per-pixel reconstruction which is separable on a modern GPU could unlock real-time capture and reconstruction in the future.
{
Combining the learned principal components of polarimetric transport and polarimetric reconstruction is another exciting direction which could facilitate robust reconstruction with fewer captures. Energy-efficient spatial probing~\cite{o2015homogeneous} is further exciting direction for future research.} 
\section{Conclusion}
\label{sec:conclusion}
We propose a computational light transport imaging method that allows, for the first time, to capture full polarization information along with temporal and spatial resolution. Our approach hinges on a novel tensor-based light transport theory that jointly models temporal, spatial and polarimetric dimensions. We analyze the low-rank polarimetric embeddings and propose a data-driven ellipsometry that learns to exploit this low-rank structure.
We instantiate our approach with two imaging systems for coaxial temporal-polarimetric imaging and spatio-polarimetric imaging.
The proposed polarimetric probing method provides in-depth analysis on material and geometry characteristics of a scene, enabling {unprecedented} decomposition of temporal-polarimetric and spatio-polarimetric light transport. {We validate the proposed method with five applications: reconstruction of shape in the presence of subsurface scattering, seeing through scattering {media}, analyzing multiple light-bounces, untangling metamerism, and analyzing crystal birefringence.} We outperform existing methods for \emph{all} applications.{We envision that the proposed method will inspire broad applications in computer graphics, vision, and beyond}.
\vspace{-3mm} 
\begin{acks}
We thank our reviewers for their invaluable comments.
We also thank Julian Knodt for helping the path-tracer implementation and Ethan Tseng for proofreading.
Felix Heide was supported by an NSF CAREER Award (2047359) and a Sony Faculty Innovation Award.
\end{acks}

\bibliographystyle{ACM-Reference-Format}
\bibliography{polar}

\end{document}